\pdfoutput=1
\documentclass[conference]{IEEEtran}      
\usepackage{graphicx} 
\usepackage{epsfig} 
\usepackage{times} 
\usepackage{amsmath} 
\usepackage{amssymb}  
\usepackage{subcaption}
\usepackage{mathtools}
\usepackage[font={small}]{caption}
\usepackage{hyperref}
\usepackage{mathrsfs}
\usepackage{amsfonts}
\usepackage{algorithm}
\usepackage{algorithmic}
\usepackage[short]{optidef}
\usepackage{balance}
\usepackage{braket}
\usepackage{pdfpages}
\usepackage{color}
\usepackage[square,sort,comma,numbers]{natbib}
\usepackage{multicol}
\usepackage{bm}
\usepackage{tikz}

\newtheorem{remark}{Remark}

\begin{document}
\title{Bridging Model-based Safety and Model-free Reinforcement Learning through System Identification of Low Dimensional Linear Models}
\author{Zhongyu Li, Jun Zeng, Akshay Thirugnanam, and Koushil Sreenath\\
\normalsize{University of California, Berkeley}\\
\normalsize{Email: \{zhongyu\_li, zengjunsjtu, akshay\_t, koushils\}@berkeley.edu}}
\maketitle

\begin{abstract}
Bridging model-based safety and model-free reinforcement learning~(RL) for dynamic robots is appealing since model-based methods are able to provide formal safety guarantees, while RL-based methods are able to exploit the robot agility by learning from the full-order system dynamics.
However, current approaches to tackle this problem are mostly restricted to simple systems. 
In this paper, we propose a new method to combine model-based safety with model-free reinforcement learning by explicitly finding a low-dimensional model of the system controlled by a RL policy and applying stability and safety guarantees on that simple model. 
We use a complex bipedal robot Cassie, which is a high dimensional nonlinear system with hybrid dynamics and underactuation, and its RL-based walking controller as an example.
We show that a low-dimensional dynamical model is sufficient to capture the dynamics of the closed-loop system.
We demonstrate that this model is linear, asymptotically stable, and is decoupled across control input in all dimensions.
We further exemplify that such linearity exists even when using different RL control policies.
Such results point out an interesting direction to understand the relationship between RL and optimal control: whether RL tends to linearize the nonlinear system during training in some cases. 
Furthermore, we illustrate that the found linear model is able to provide guarantees by safety-critical optimal control framework, \textit{e.g.}, Model Predictive Control with Control Barrier Functions, on an example of autonomous navigation using Cassie while taking advantage of the agility provided by the RL-based controller.
\end{abstract}

\section{Introduction}
It is challenging for robotic systems to achieve control objectives with coupled dynamical models while considering input, state and safety constraints.
Model-based safety-critical optimal control methods that combine control barrier functions (CBFs)~\cite{ames2019control} or  Hamilton-Jacobi~(HJ) reachability analysis~\cite{bansal2017hamilton} are able to provide formal safety guarantees for control, planning and navigation problems on different platforms such as autonomous driving~\cite{ames2014control, son2019safety, he2021rule}, aerial systems~\cite{chen2018hamilton, singletary2020safety, chen2021fastrack} and legged robots~\cite{choi2020reinforcement, grandia2021multi}.
However, previous work on applying such safety-critical methods are only able to be validated on relatively simple or low-dimensional systems such as a 5 Degree-of-Freedom~(DoF) $2D$ bipedal robot~\cite{nguyen2018dynamic} or a quadrotor with decoupled dynamics~\cite{herbert2017fastrack}.
When encountering the safety-critical control problem on complex and high-dimensional systems, such as a $3D$ bipedal robot Cassie~\cite{li2020animated} which has $20$ DoF and hybrid walking dynamics, model-based methods will face challenges since the full-order dynamics model of the robot is computationally not tractable for online implementation with current computing tools.

\begin{figure}[t]
    \centering
    \includegraphics[width=0.8\linewidth]{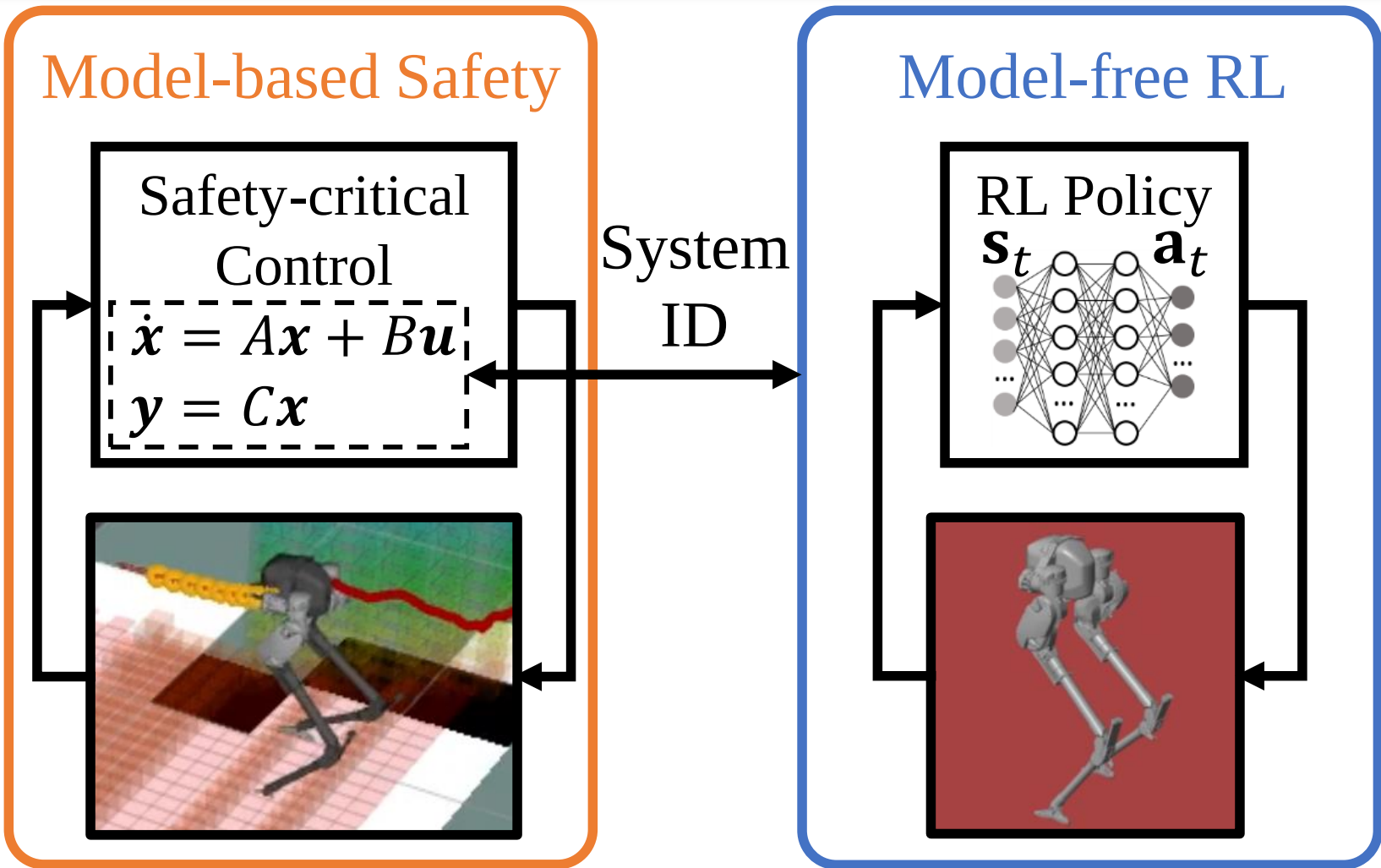}
    \caption{Our proposed method to bridge model-based safety-critical control and model-free reinforcement learning~(RL). This method applies system identification of a closed-loop system which is a robot controlled by its RL policy in order to find its low-dimensional models. Once an explicit model is obtained, ideally, a linear model, model-based controllers can be applied on this model to provide safety guarantees on the robot controlled by its RL-based policy. This method is able to exploit the advantages of RL to utilize full-order dynamics of a system and model-based methods to have stability and safety guarantees. Thus, our method is more applicable in the real-world deployment on complex systems such as legged robots.}
    \label{fig:main} 
\end{figure}

Model-free reinforcement learning~(RL) methods, on the other hand, are able to leverage full-order dynamics model of the robot during offline training in simulation to provide a policy for online control.
With the recent progress on solving the sim-to-real problem, model-free RL demonstrates the capacity to control a large range of dynamic robots in the real world~\cite{hwangbo2017control,peng2018sim,xie2018feedback,lee2020learning,peng2020learning,siekmann2021sim,li2021reinforcement,castillo2021robust}. 
For example, with the help of model-free RL, a robust and versatile walking controller for a bipedal robot Cassie is obtained in~\cite{li2021reinforcement} to reliably track given commands of walking velocities, walking height, and turning yaw rate in experiments on the hardware.
This RL-based controller provides significant improvements over traditional model-based walking controllers~\cite{gong2019feedback, li2020animated} by showing larger feasible commands and the ability to stay robust to random perturbations.

Such robust controllers create interesting questions in the community: where does the robustness arise from and whether we can formally assess the stability of such a RL-based controller/policy?
Addressing such questions is very interesting as studying properties of a RL policy can further our understanding of the reason that RL demonstrates advantages on controlling highly dynamic systems.
Moreover, if we can find an explicit dynamic model of a system controlled by RL, we may be able to utilize such a model to formally guarantee safety for such autonomous systems.
However, this is challenging as the policy obtained by RL is usually represented by a high dimensional nonlinear neural network and explicit analysis on nonlinear systems with RL policies with rigorous proofs is still an unsolved problem.
In this paper, we seek to ascertain the feasibility to find and study a low dimensional explicit model of a complex dynamic robot driven by a RL policy and to utilize such a simple model to provide guarantees on stability and safety during safety-critical tasks, such as autonomous navigation, as abstractly illustrated in Fig.~\ref{fig:main}.

\subsection{Related Work}

\subsubsection{Safety \& Learning}
There has been some exciting progress to bridge safety and learning. 
Previous approaches can be summarized into three classes: (a) learning dynamics in the model-based control setting, (b) increasing robustness for RL by model-based safety, and (c) providing learned stability and safety using existing model-based controllers.
We provide more details on each of these approaches next.

\paragraph{Learning Dynamics in Model-based Control}
Safety can be guaranteed based on modern control frameworks. 
One approach is to integrate adaptive control with standard machine learning methods, such as NN~\cite{vinitsky2020robust}, GP~\cite{gahlawat2020l1} and DNN~\cite{joshi2019deep,manek2020learning,chou2021model}.
The safety properties are usually considered on the whole system with some parts being learned in MPC problems~\cite{bujarbaruah2018adaptive, kohler2021robust}, shielding~\cite{alshiekh2018safe}, Control Barrier Functions~(CBFs)~\cite{cheng2019end}, Hamiltonian analysis~\cite{bansal2021deepreach}.
These approaches can guarantee safety for tasks such as stabilization~\cite{kamthe2018data, westenbroek2020learning} and tracking~\cite{ostafew2016robust, fan2020deep}. 
However, these approaches usually make assumptions to apply on a known model structure, such as control affine or linear system with bounded uncertainty, which becomes challenging to apply on high-dimensional nonlinear systems.

\paragraph{Model-based Safety in Reinforcement Learning}
Another approach to bridge safety and learning is to increase robustness and safety in RL tasks by model-based methods.
When using RL to solve a control problem through trial and error, safety is considered by imposing input constraints or safety rewards.
Safe exploration~\cite{moldovan2012safe} and safe optimization~\cite{sui2015safe} of MDPs under unknown or selected cost functions can be formulated.
Constrained MDPs are also common in RL tasks with Lagrangian methods~\cite{chow2017risk} and generalized Lyapunov/barrier functions~\cite{chow2018lyapunov, cheng2019end, dawson2022safe} and shielding~\cite{alshiekh2018safe}. 
Nonetheless much of the work remains confined to rather naive simulated tasks, such as moving a $2D$ agent on a grid map.

\paragraph{Learned Certifications for Model-based Controllers}
Remaining methods for connecting safety and learning are to provide certification on learned stability and constraint set using existing model-based controllers.
Stability criterion can be achieved by Lyapunov analysis on region of attraction (RoA)~\cite{richards2018lyapunov, dai2021lyapunov,boffi2020learning} or by Lipschitz-based safety~\cite{jin2020stability} during training.
Safety with constraint set certifications can be learned using control synthesis such as feedback linearization controllers~\cite{westenbroek2020feedback}, CBFs~\cite{saveriano2019learning, yaghoubi2020training, taylor2020learning, robey2020learning, marvi2021safe, robey2021learning, wang2021learning}, and Hamilton-Jacobi~(HJ) reachability analysis~\cite{fisac2019bridging, ivanovic2019barc, bansal2021deepreach}.
However, all these approaches are only validated on simple dynamic systems such as $5$~DoF bipedal robot in simulation or $7$~DoF static robot arm in the real world.
This is because these approaches usually suffer from a curse of dimensionality, in other words, finding a valid control barrier function~\cite{ames2014control} or a backward reachable set~\cite{bansal2017hamilton} for high dimensional systems remain unsolved problems.
As we will see, the proposed method in this paper is an “inverse" of such a methodology: we use model-based methods to find certifications on closed-loop systems with RL-based controllers, which is more practical to apply on high order and nonlinear systems. 

\subsubsection{Low-dimensional Structure of Deep Learning}
Finding low-dimensional structures of high-dimensional data is widely used in the statistical learning field, such as PCA, kernels, etc~\cite{wright2021high}.
More recently, researchers in the deep learning field realize that learning on nonlinear functions in high-dimensional space may tend to linearize and compress the system and obtain a low-dimensional linear representation of it within the learning components~\cite{karl2016deep, mania2018simple, allen2019convergence, wei2020implicit, chan2021redunet}. 
But this is still an open question and under debate. 
In the reinforcement learning domain, model-based RL methods usually choose to learn local dynamics models, sometimes by fitting linear models, and utilize the learned models to design optimal control policy~\cite{watter2015embed,levine2013guided,levine2016end}. 
However, the control performance using the model-based RL relies on the learned reduced-order model and therefore cannot exploit robot's full-order dynamics. 
Model-free RL, as a counterpart, does not require to explicitly utilize a model to develop the control policy, and shows advantages on controlling high-order complex robots by leveraging the robot's full order dynamics implicitly through samples~\cite{hwangbo2017control,peng2018sim,li2021reinforcement}. 
However, there is little effort exerted on finding the low-dimensional structure of a system driven by model-free RL as the optimization of the model-free RL is usually realized through trial and error.
In this work, we show some evidence that model-free RL may learn through linearizing the nonlinear system in some cases, such as tracking control demonstrated in this paper. 

\subsection{Contributions}
In this paper, we propose a new direction to bridge model-based safety and model-free reinforcement learning by system identifying a low dimensional model of a closed-loop system controlled by a learned policy, as shown in Fig.~\ref{fig:main}. 
We use the walking controller of a life-sized bipedal robot Cassie, which is represented by a deep neural network optimized by reinforcement learning in~\cite{li2021reinforcement}.
This is one of the first attempts to apply system identification using a low-dimensional model on the complex system controlled by a RL policy.
We find out that a linear model is sufficient to describe this closed-loop system.
This linear model shows desirable properties such as all dimensions are decoupled, minimum phase, and asymptotically stability.
We further show that linearity exists across multiple RL policies under a provided criterion.
Such a method shows the possibility to answer the questions from the control community to the learning community: how to analyze the stability of the policies obtained by RL.
Additionally, we demonstrate that the linearity analysis can reflect the convergence of RL. 
Furthermore, based on the found linear model, we propose one of the first safe navigation framework using a high-dimensional nonlinear robot, a bipedal robot Cassie, that combines low-level RL policy for locomotion control and high-level safety-critical optimal control for navigation. 
This paper serves as an introduction for a new perspective to understand the relationship between RL and optimal control, and to practically utilize model-free RL for safety critical control on complex systems.  
\label{sec:intro}

\section{Methodology}
In this section, the experimental platform, the bipedal robot Cassie, and its RL policy for locomotion controller are briefly introduced. 
Moreover, the method utilized to find the low-dimensional model by system identification through input-output pairs is also presented. 

\subsection{Cassie and its RL policy for Walking Control}

\begin{figure}
    \centering
    \includegraphics[width=\linewidth]{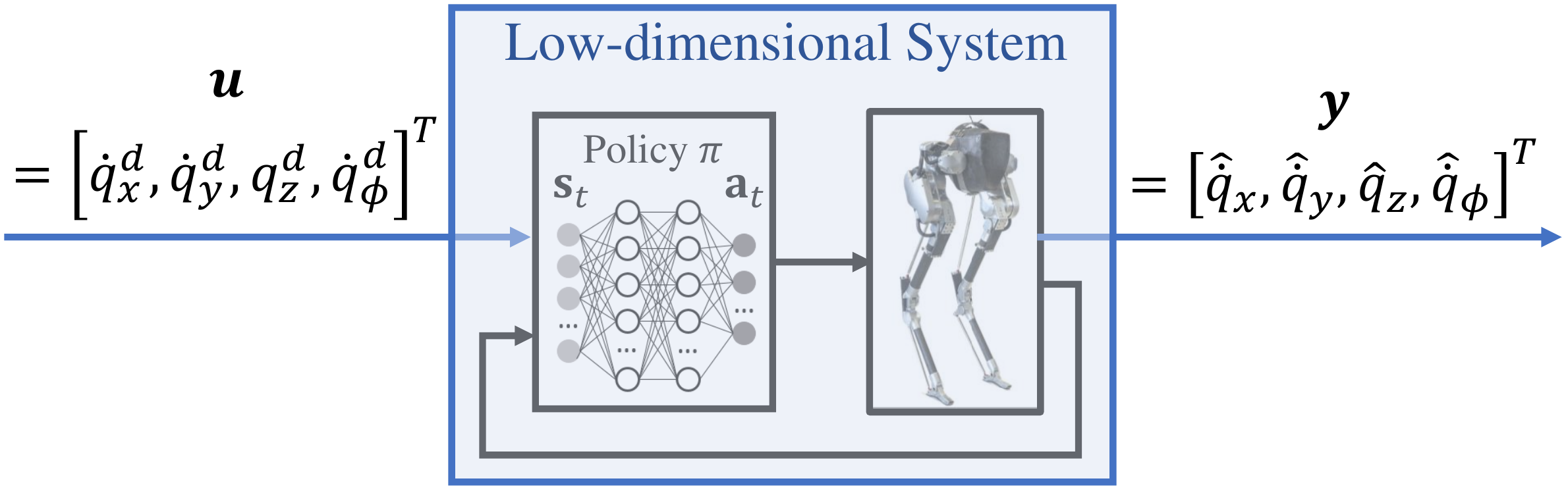}
    \caption{System identification on the closed-loop system. The low-dimensional system to identify is Cassie, a high-dimensional nonlinear system, controlled by its RL policy. The input $\mathbf{u}$ of this simple system is desired commands including sagittal walking speed $\dot{q}^d_x$, lateral walking speed $\dot{q}^d_y$, walking height $q_z$, and turning yaw speed $\dot{q}^d_\phi$. The output $\mathbf{y}$ is the robot's measured output values driven by the low-level RL policy.}
    \label{fig:sys_id_intro} 
\end{figure}

Cassie is a life-size bipedal robot. It has 10 motors and 4 underactuated joints connected by springs. A detailed introduction of Cassie can be found in~\cite{gong2019feedback,li2020animated}. 
There has been some exciting progress on applying model-free reinforcement learning to obtain locomotion controllers on Cassie in the real world~\cite{xie2018feedback,li2021reinforcement,siekmann2021sim}.
Among these work, \cite{li2021reinforcement} develops a robust and versatile walking controller on Cassie that can track variable commands via RL. 
This policy is represented by a 2-layer fully-connected neural network with $512$ $tanh$ nonlinear activations in each layer, and directly outputs 10 dimensional desired motor positions which are then used in a joint-level PD controller to generate motor torques in real time.
The policy observation includes reference motion to imitate, robot current states, and $4$ timesteps past robot states and actions. 
This single policy is able to track a $4$ dimensional command, which is comprised of the desired sagittal walking speed $\dot{q}^d_x$, lateral walking speed $\dot{q}^d_y$, walking height $q^d_z$, and turning yaw rate $\dot{q}^d_\phi$.
The agent is trained in Mujoco~\cite{todorov2012mujoco} which is a physics simulator, and the policy is optimized by Proximal Policy Optimization~(PPO)~\cite{schulman2017proximal}.

In real-world experiments, the policy demonstrates considerable robustness and sophisticated recoveries.
For example, in a recovery case demonstrated in~\cite{li2021reinforcement}, the robot almost falls down and deviates a lot from the nominal walking states, but the controller is still able to regulate the system without letting the states go unbounded and triggering instability. 
Although it is very hard to show the explicit stability and RoA of a controller represented by a high dimensional nonlinear neural network, 
it is possible that the robustness of this RL-based controller is due to its stability while having a relatively large RoA, or ideally, being globally stable.

Therefore, in this work, we utilize two types of RL-based controllers based on~\cite{li2021reinforcement}: (1) the same multi-layer perceptrons~(MLP) introduced in~\cite{li2021reinforcement}, (2) the same MLP but with an additional encoder represented by a 2-layer convolutional neural network~(CNN) to record longer history of robot states and actions that last 2 seconds (66 timesteps). 


\begin{figure}
\centering
\begin{subfigure}{\linewidth}
  \centering
  \includegraphics[width=\linewidth]{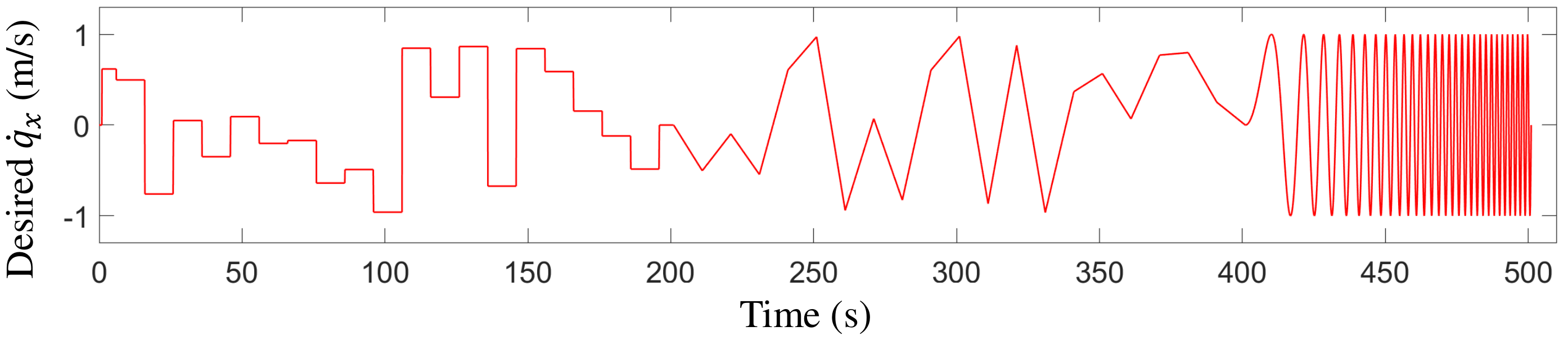}
  \caption{An Example of System Inputs: Desired Sagittal Walking Velocity}
  \label{subfig:example-input-vx}
\end{subfigure}
\begin{subfigure}{\linewidth}
  \centering
  \includegraphics[width=\linewidth]{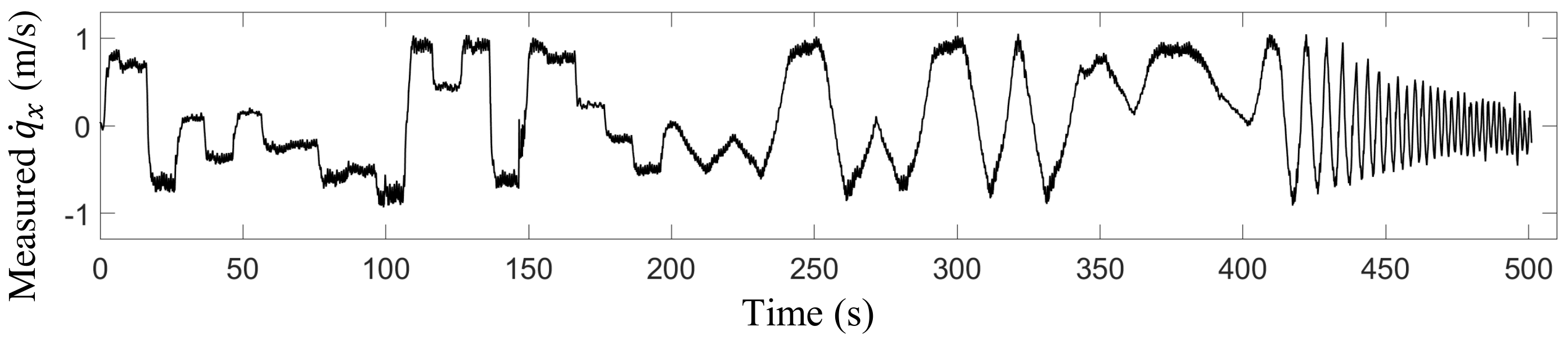}
  \caption{System Outputs using MLP: Measured Sagittal Walking Velocity}
  \label{subfig:example-output-vx}
\end{subfigure}
\begin{subfigure}{\linewidth}
  \centering
  \includegraphics[width=\linewidth]{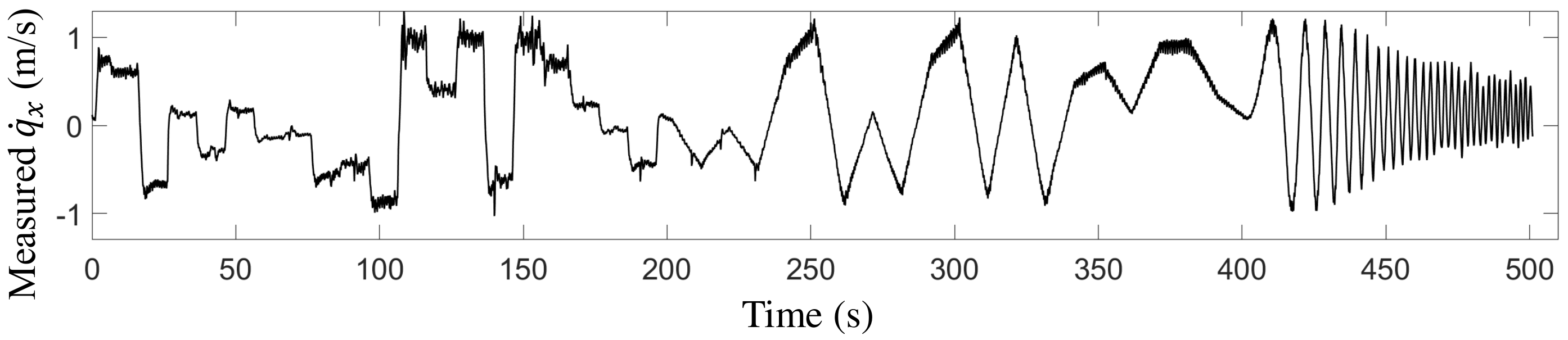}
  \caption{System Outputs using CNN: Measured Sagittal Walking Velocity}
  \label{subfig:example-output-vx-cnn}
\end{subfigure}
\caption{An example of input-output pair used in this paper to identify the dynamics of Cassie being controlled by the RL-based walking controllers. (a) The input signal which is the desired sagittal waking velocity $\dot{q}^d_x$. The input command consists of steps, ramps and chirp signals. (b) The measured Cassie sagittal walking velocity $\hat{\dot{q}}_x$ which are the step response in the first $200$~s, ramp response from $200$~s to $400$~s, and frequency response in the last $100$~s, using the MLP policy. (c) System response using the CNN policy.}
\label{fig:example-io-vx}
\end{figure}

\subsection{System Identification}
In order to understand the dynamics of the RL-based controller, we choose to identify the entire closed-loop system comprising of Cassie being controlled by the walking controllers obtained through model-free RL. 
The input $\mathbf{u}$ to this closed-loop system has only $4$ dimensions, which is the control reference $[\dot{q}^d_x, \dot{q}^d_y, q^d_z, \dot{q}^d_\phi]^T$. 
The output of this system is the observed robot walking velocities and walking height, i.e. $\mathbf{y} = [\hat{\dot{q}}_x, \hat{\dot{q}}_x, \hat{q}_z, \hat{\dot{q}}_\phi]^T$. 
We then collect input-output pairs of this system in a high-fidelity simulator of Cassie built on MATLAB Simulink.
Please note that the control policy is trained on Mujoco and has no access to the data from Simulink during training and has also been shown to work in Simulink.

One example of the input-output signal of the sagittal walking velocity dimension is shown in Fig.~\ref{fig:example-io-vx}.
There are three types of the input signals: 1) random step signal as shown in $0$-$200$~s in Fig.~\ref{subfig:example-input-vx}, 2) random ramp signal as demonstrated in $200$-$400$~s in Fig.~\ref{subfig:example-input-vx}, and 3) swept frequency sine wave~(\textit{chirp}) whose frequency linearly expands from $0$~Hz to $1$~Hz in $400$-$500$~s in Fig.~\ref{subfig:example-input-vx}.
The resulting robot outputs represent the step response, ramp response, and frequency response of this system, respectively, as illustrated in the corresponding time span in Figs.~\ref{subfig:example-output-vx} and \ref{subfig:example-output-vx-cnn}. 
After selecting a model structure, the parameters of the model can be fitted by combining the input signals and the output signals measured from Cassie driven by the RL-based controller. 
As it is unsafe for the life-sized bipedal robot to experimentally follow those random input signals in real-life, Simulink provides a safe validation domain in this paper.

\begin{remark}
We note walking robots are hybrid systems.  
Identifying the closed-loop model of a walking robot between successive steps in an event-based manner provides us a discrete model representing the step-to-step transitions on a chosen Poincare' section.
In contrast, here we identify a model based on time-series data to obtain a continuous-time input-output model.
In both of the above cases, we get past the need to explicitly identify the hybrid dynamics of the walking robot.
\end{remark}

\label{sec:framework}

\section{Linearity}
\label{sec:linearity}

In this section, we use a linear model structure, \textit{i.e.}, $\dot{\mathbf{x}} = A\mathbf{x} + B\mathbf{u}$, $\mathbf{y}=C\mathbf{x}$, to fit the input-output pairs of the closed-loop system which is Cassie being controlled by the RL-based policy. 
This system, as shown in Fig.~\ref{fig:sys_id_intro}, has four inputs, $\mathbf{u} = [\dot{q}^d_x, \dot{q}^d_y, q^d_z, \dot{q}^d_\phi]^T \in \mathbb{R}^4$ and four outputs, $\mathbf{y} = [\hat{\dot{q}}_x, \hat{\dot{q}}_y, \hat{q}_z, \hat{\dot{q}}_\phi]^T \in \mathbb{R}^4$. 
Linear models show reasonably good fitting accuracy at each of four dimensions while other dimensions have constant inputs. 
These four dimensions are sagittal walking velocity $\hat{\dot{q}}_x=f_{\dot{x}}(\mathbf{x}_{\dot{x}},{u}_{\dot{x}})$, lateral walking velocity $\hat{\dot{q}}_y=f_{\dot{y}}(\mathbf{x}_{\dot{y}},{u}_{\dot{y}})$, walking height $\hat{q}_z=f_{z}(\mathbf{x}_{z},{u}_{z})$, and turning yaw rate $\hat{\dot{q}}_\phi=f_{\dot{\phi}}(\mathbf{x}_{\dot{\phi}},{u}_{\dot{\phi}})$.
The fitted linear systems show that they are stable systems. 
Moreover, the results also show that the input frequency should stay below certain threshold to preserve the linearity of the system. 

\subsection{Identified Linear Systems}\label{subsec:fiting}

\begin{figure}
\centering
\begin{subfigure}{0.48\linewidth}
  \centering
  \includegraphics[width=\linewidth]{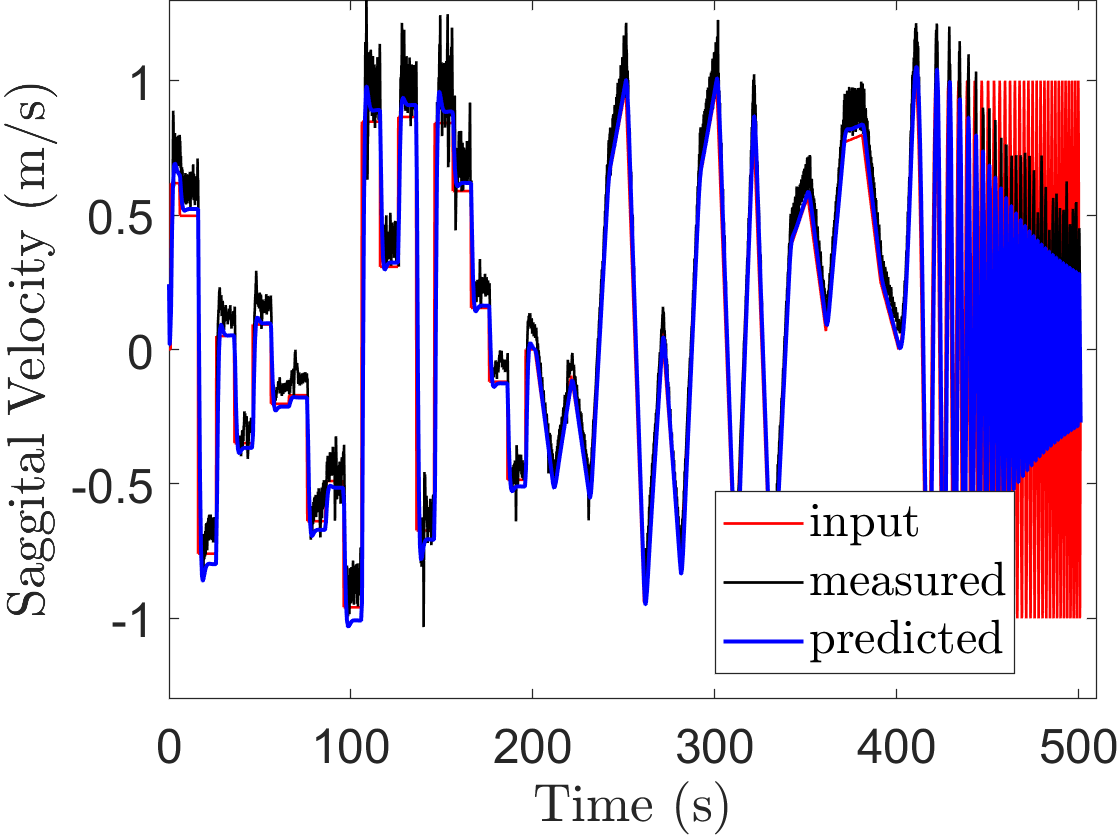}
  \caption{Fitted Linear Model of $\dot{q}_x$}
  \label{subfig:vx_id}
\end{subfigure}
\begin{subfigure}{0.48\linewidth}
  \centering
  \includegraphics[width=\linewidth]{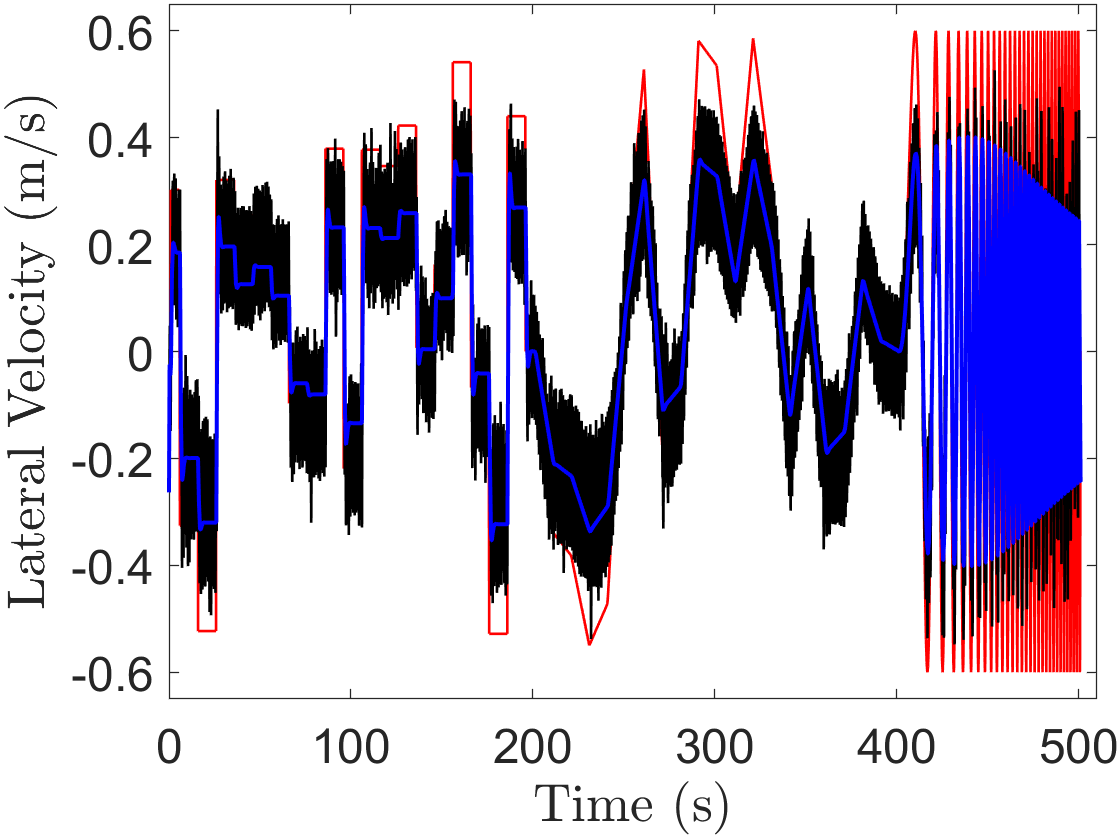}
  \caption{Fitted Linear Model of $\dot{q}_y$}
  \label{subfig:vy_id}
\end{subfigure}
\begin{subfigure}{0.48\linewidth}
  \centering
  \includegraphics[width=\linewidth]{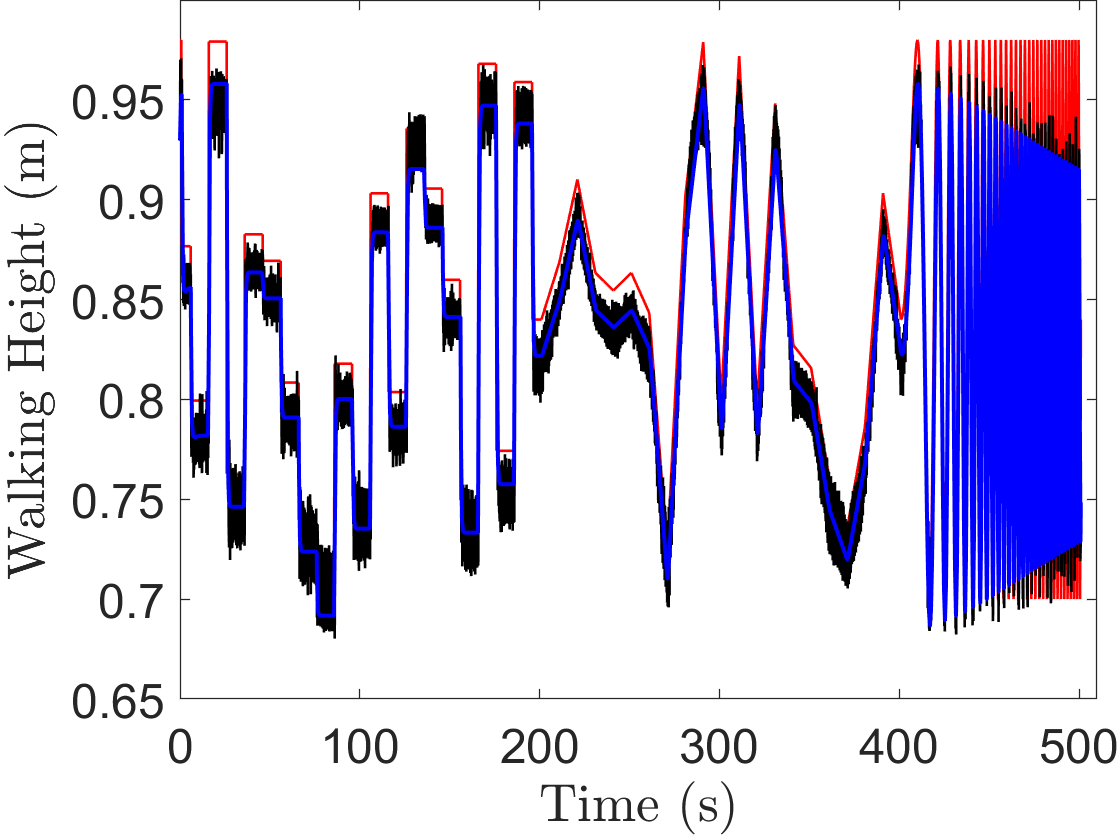}
  \caption{Fitted Linear Model of $q_z$}
  \label{subfig:wh_id}
\end{subfigure}
\begin{subfigure}{0.48\linewidth}
  \centering
  \includegraphics[width=\linewidth]{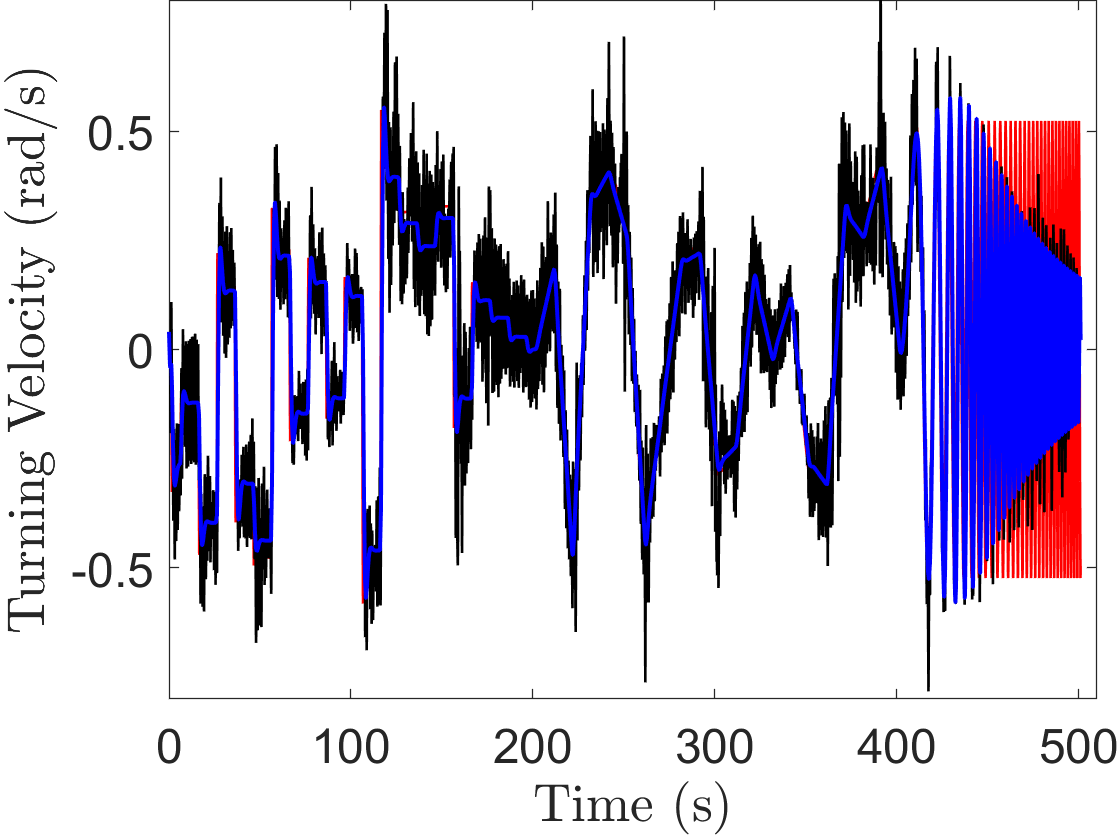}
  \caption{Fitted Linear Model of $\dot{q}_\phi$}
  \label{subfig:vyaw_id}
\end{subfigure}
\caption{Fitting results using linear models for all four dimensions of Cassie being controlled by a CNN-based RL policy. The red lines are the random step, ramp, and chirp input signals, the black lines represent measured Cassie measured walking velocities and walking height. The blue lines are the predicted result using a fitted linear model. This prediction result uses one time-step measurement to predict next $5$ time step system output using the input and model. The frequency of the all the measured input-output data is $2$~kHz.}
\label{fig:id_result}
\end{figure}

The fitting results using linear models for the four dimension outputs of the system which is Cassie being controlled by the CNN-based RL policy for walking controller are demonstrated in Fig.~\ref{fig:id_result}.
The fitting results of MLP-based RL walking controller are attached in Appendix~\ref{subsec:fitting-results-mlp}.
During the fitting phase of each dimension, the measured input-output data of the system are used to find the system parameters of the selected linear model in order to minimize the difference between the predicted and measured system outputs.
The structure of the linear model, \textit{i.e.}, numbers of zeros and poles, are searched in order to maximize fitting accuracy while having the simplest structure which has the least number of zeros and poles by looking at Hankel singular values of the fitted system~\cite{ljung1998system}.
For different dimensions of the system, the dynamic model may be varied. 
During the validation phase, the fitted model is utilized to predict next $5$ step system output using the system input and $1$ step measured output data, the prediction results are shown as the blue lines in Fig.~\ref{fig:id_result}. 

The fitted dynamics of sagittal walking velocity $f_{\dot{x}}(\mathbf{x}_{\dot{x}},{u}_{\dot{x}})$ is given by:
\noindent
\begin{equation}
    Y_{\dot{x}}(s) = \frac{0.4694s^2 + 6.089s+8.697}{s^3+6.432s^2+11.03s+8.274} U_{\dot{x}}(s)
\label{eq:vx_dynamics}
\end{equation}
\noindent
with a prediction accuracy \footnote{The \textit{prediction accuracy} is termed as fit percentage which is obtained by $(1-\text{NRMSE})\times100$ where NRMSE is the Normalized Root Mean Squared Error between the predicted output $\hat{a}$ and actual output $a$ calculated via $||a-\hat{a}||_2/||a-\text{mean}(a)||_2$.} of $79.67\%$ when compared with the ground truth measured data, as shown in Fig.~\ref{subfig:vx_id}. 
$Y(s)$ and $U(s)$ are the system output and input in $s$-domain after Laplace transform of the system dynamics equation in time domain, respectively.

The dynamics of the lateral walking velocity $f_{\dot{y}}(\mathbf{x}_{\dot{y}},{u}_{\dot{y}})$ is obtained by using a linear system of $3$ poles and $1$ zeros, and it can be written as:
\noindent
\begin{equation}
Y_{\dot{y}(s)} =
\frac{13.59s+24.56}{s^3+11.48s^2+32.5s+40.13}{U_{\dot{y}}(s)}
\label{eq:vy_dynamics}
\end{equation}
\noindent
which has a prediction accuracy of $55.87\%$. 
However, as shown in Fig.~\ref{subfig:vy_id}, there is a large oscillation of the robot base lateral direction when Cassie is walking. After applying a low-pass filter with a cut-off frequency of $5$~Hz on the measured lateral velocity, the prediction accuracy on the flitted signal is $83.03\%$ using the same fitted model via ~\eqref{eq:vy_dynamics}.

The dynamics of walking height $f_{z}(\mathbf{x}_{z},{u}_{z})$ and the dynamics of turning yaw velocity $f_{\dot{\phi}}(\mathbf{x}_{\dot{\phi}},{u}_{\dot{\phi}})$ are fitted as:
\noindent
\begin{equation}
    Y_{z}(s) = \frac{145.9s + 37.55}{s^3+46.43s^2+161s+38.38} U_{z}(s)
\label{eq:wh_dynamics}
\end{equation}
\noindent
and 
\noindent
\begin{equation}
    Y_{\dot{\phi}}(s) = \frac{0.3078s^2 + 5.267s +  5.553}{s^3+4.595s^2+7.528s+6.045} U_{\dot{\phi}}(s)
\label{eq:vyaw_dynamics}
\end{equation}
\noindent 
and with prediction accuracy of $85.67\%$ in Fig.~\ref{subfig:wh_id} and $65.58\%$ ($81.77\%$ after filtering the measured output) in Fig.~\ref{subfig:vyaw_id}, respectively.  

According to the prediction performance as demonstrated in Fig.~\ref{fig:id_result}, all of the four dimensions show reasonably good linearity as the fitting accuracy of them are all around $80\%$.     
Therefore, we can come to a conclusion that all of these $4$ dimensions of Cassie during walking shows linearity when the robot is controlled by the RL policy.  

\subsection{Stability}

Stability analysis can be applied on each dimension after obtaining the low-dimensional linear dynamics model of the closed-loop system.
Since the dynamics of each dimension is able to be explicitly expressed by a transfer function, a commonly used stability criterion is the position of the poles of the system. 
Please note that during the model identification in Sec.~\ref{subsec:fiting}, we allow the algorithm to find unstable models as long as the fitting results are better than stable models, \textit{i.e.}, we don't impose stable model constraint during fitting.

\begin{figure}
\centering
\begin{subfigure}{0.425\linewidth} 
  \centering
  \includegraphics[width=\linewidth]{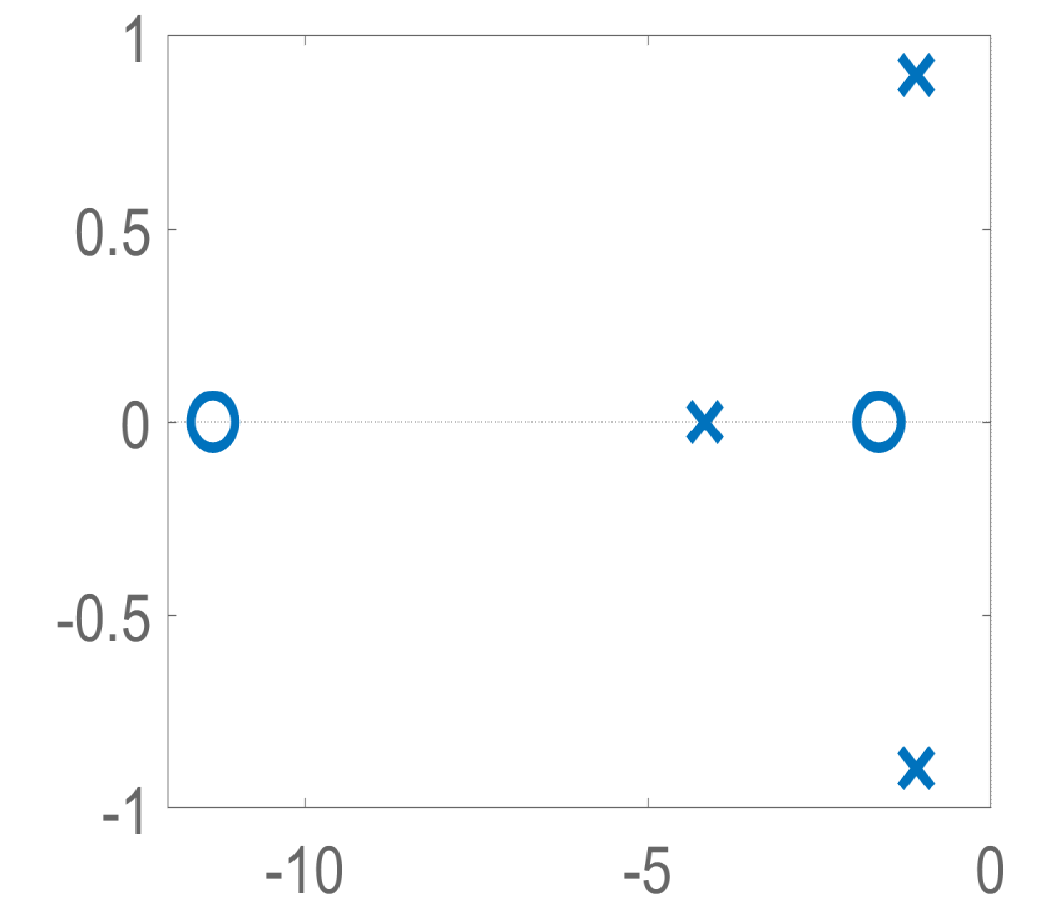}
  \caption{Pole-Zero Plot of $\dot{q}_x$}
  \label{subfig:vx_zp}
\end{subfigure}
\begin{subfigure}{0.425\linewidth}
  \centering
  \includegraphics[width=\linewidth]{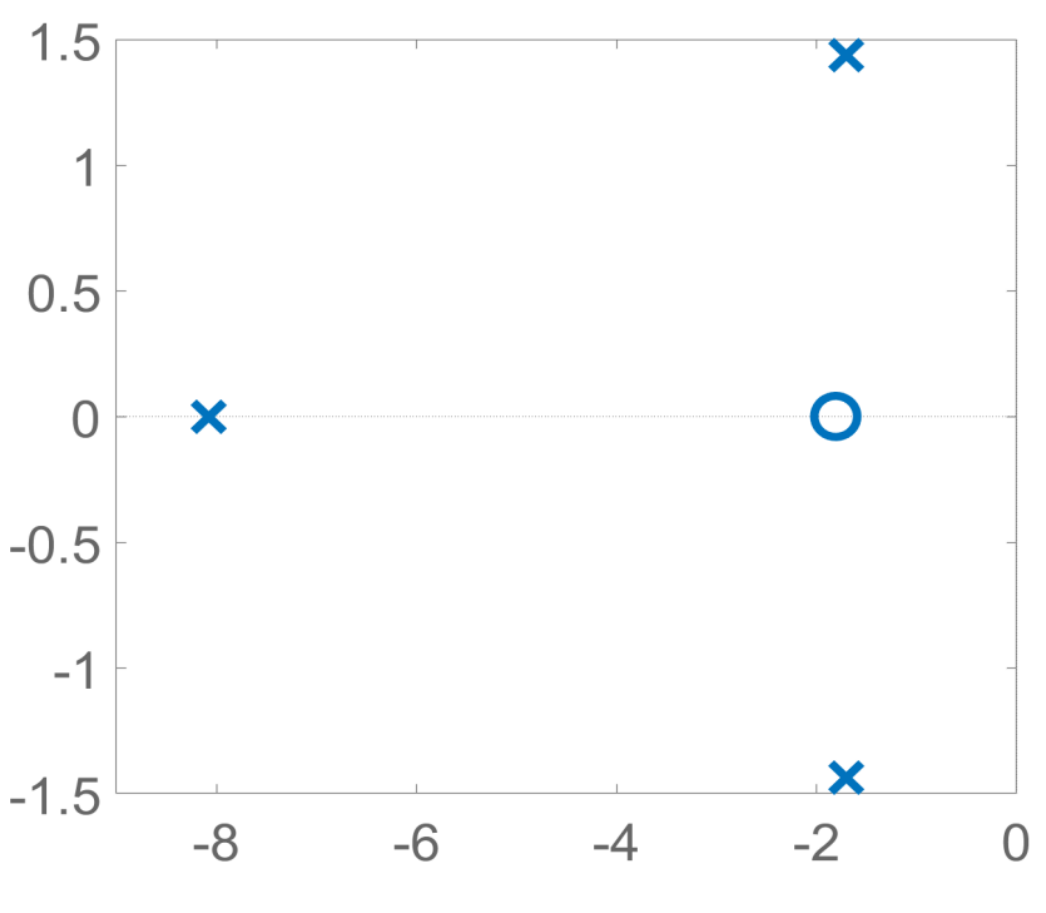}
  \caption{Pole-Zero Plot of $\dot{q}_y$}
  \label{subfig:vy_zp}
\end{subfigure}
\begin{subfigure}{0.425\linewidth}
  \centering
  \includegraphics[width=\linewidth]{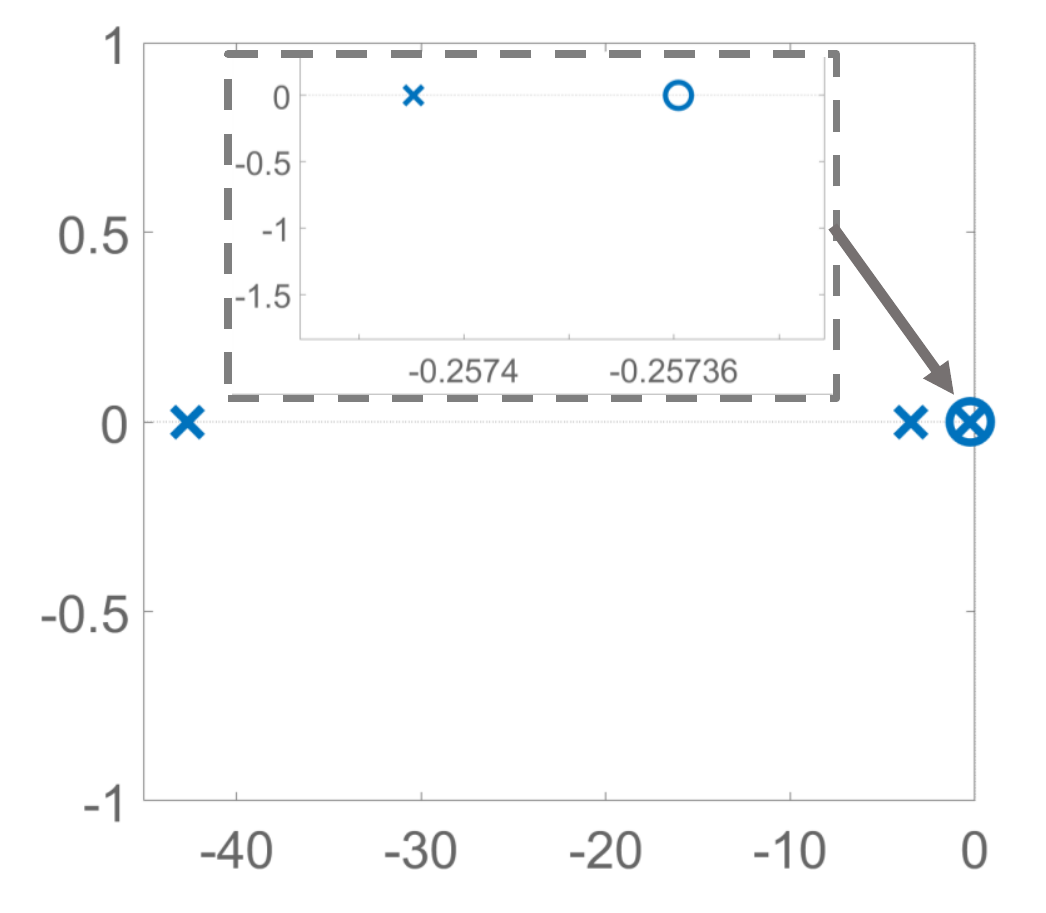}
  \caption{Pole-Zero Plot of $q_z$}
  \label{subfig:wh_zp}
\end{subfigure}
\begin{subfigure}{0.425\linewidth}
  \centering
  \includegraphics[width=\linewidth]{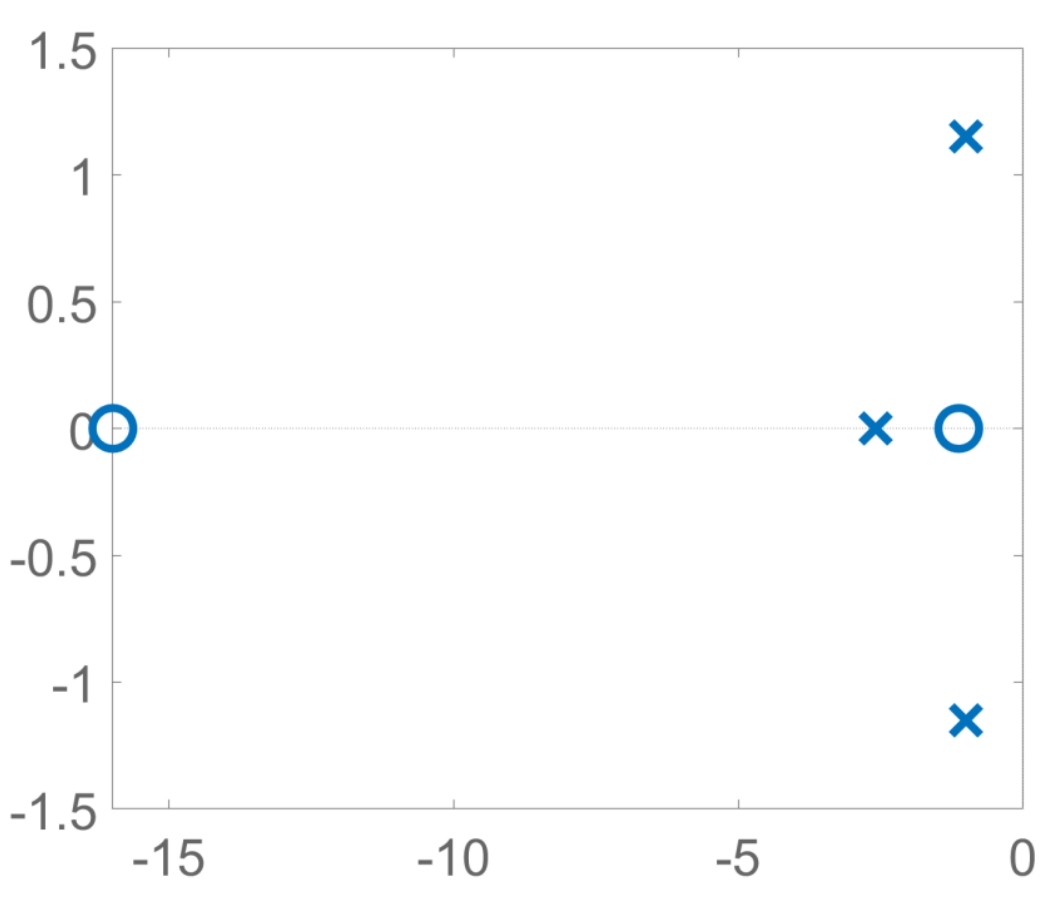}
  \caption{Pole-Zero Plot of $\dot{q}_\phi$}
  \label{subfig:vyaw_zp}
\end{subfigure}
\caption{Pole-Zero plots of the identified linear systems that uses CNN policy for sagittal walking velocity, lateral walking velocity, walking height, and turning yaw rate, respectively. All poles and zeros are on the left hand plane~(LHP).}
\label{fig:zero-pole}
\end{figure}

The plots of poles and zeros of all four dimension of the system using the CNN policy are shown in Fig.~\ref{fig:zero-pole}.
As illustrated in Fig.~\ref{subfig:wh_zp}, there is a very close pole-zero pair of the identified $q_z$ dynamics. 
However, that pole-zero pair cannot be cancelled otherwise it will lead to worse fitting accuracy. 

According to the plot, all of the identified linear systems are asymptotically stable as all the poles are on the Left Half Plane~(LHP)~\cite{franklin2002feedback}.
Moreover, all zeros are on the LHP as well, therefore, all dimensions are minimum phase.
This shows that the system controlled by CNN-based RL policy has a very nice property that the closed-loop system with RL-based policy can be represented through a linear model which is also stable.
Such stability provides us some insights regarding the robustness of the RL policy demonstrated in the real world~\cite{li2021reinforcement}.

\begin{remark}
All poles of the linear system obtained through system identification of the closed-loop system under MLP-based RL policy (pole-zero plots are in Appendix \ref{subsec:pole-zero-plot-mlp}) are asymptotically stable, however, the system driven by MLP is non-minimum phase in dimensions of the sagittal and lateral walking velocity. 
Therefore, when choosing to utilize the robot driven by such an RL policy, the high-level controller/planner should be carefully designed.
Such stability analysis provides a guidance to utilize the system controlled by RL policies.
\end{remark}

\subsection{Criterion for Linearity} \label{subsec:criterion}
Although Cassie driven by the RL policy as a walking controller shows reasonably good linearity as shown in Fig~\ref{fig:id_result}, there is a criterion for the existence of linearity. 
That is, the input frequency cannot exceed a threshold which is around $f_c = 0.6$~Hz. 
The frequency $0.6$~Hz is an empiric value and this is found when the input tends to excite nonlinearity of the system if the input frequency exceeds that value during system identification. 
The fitted linear model shows worse fitting accuracy if the chirp signal enters the region where frequency is larger than $0.6$ Hz in Fig.~\ref{fig:id_result}. 
This may be related to the stepping frequency of Cassie. 
Given the fact that Cassie steps on each feet at a stepping rate of $1.25$~Hz, $f_c = 0.6$~Hz is one half of it. 
The existence of this cutoff frequency may be due to fact that the RL policy is unable to change the walking velocities, such as $\dot{q}_x$, $\dot{q}_y$, $\dot{q}_\phi$, faster than the time Cassie takes to complete one walking step (comprising of two steps: left foot stance followed by right foot stance). 
Another evidence of this validity of this relationship is that there is no obvious linearity lost after this frequency threshold in the identified $q_z$ dynamics. This is because changing the walking height doesn't need to wait until completion of one walking step, \textit{e.g.}, robot can change the walking height by changing the length of the stance leg at any time. 
The phenomenon of loosing linearity after this frequency threshold will frequently appear in the later part of this paper.

\section{Decoupled System}
\begin{figure*}[!htp]
\centering
\begin{subfigure}{0.115\linewidth}
  \centering
  \includegraphics[width=\linewidth]{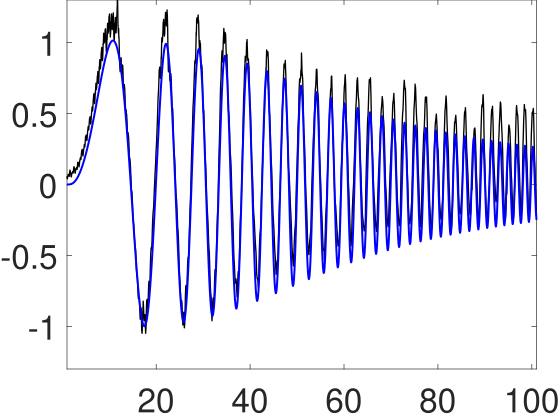}
  \caption{Fitting: chirp $\dot{q}_x$, constant rests}
  \label{subfig:dc_vx_only}
\end{subfigure}
\begin{subfigure}{0.115\linewidth}
  \centering
  \includegraphics[width=\linewidth]{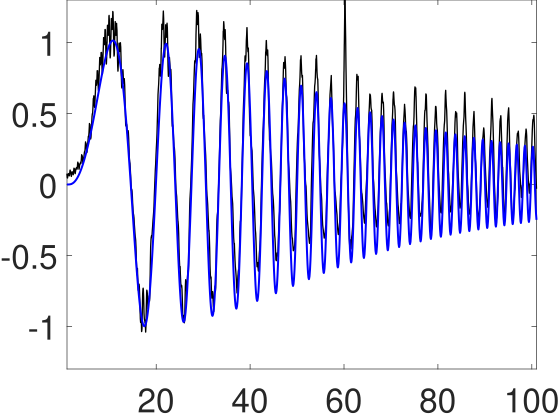}
  \caption{Testing input: chirp $\dot{q}_x$, chirp $\dot{q}_y$}
  \label{subfig:dc_vx_vychirp}
\end{subfigure}
\begin{subfigure}{0.115\linewidth}
  \centering
  \includegraphics[width=\linewidth]{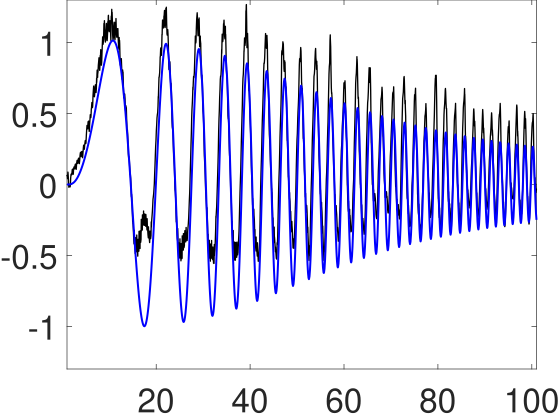}
  \caption{Testing input: chirp $\dot{q}_x$, chirp $q_z$}
  \label{subfig:dc_vx_whchirp}
\end{subfigure}
\begin{subfigure}{0.115\linewidth}
  \centering
  \includegraphics[width=\linewidth]{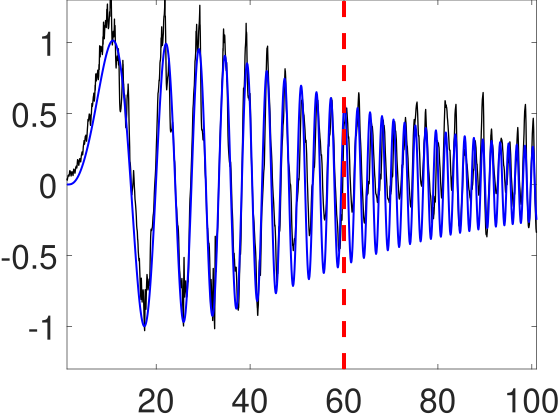}
  \caption{Testing input: chirp $\dot{q}_x$, chirp $\dot{q}_\phi$}
  \label{subfig:dc_vx_vyawchirp}
\end{subfigure}
\begin{subfigure}{0.115\linewidth}
  \centering
  \includegraphics[width=\linewidth]{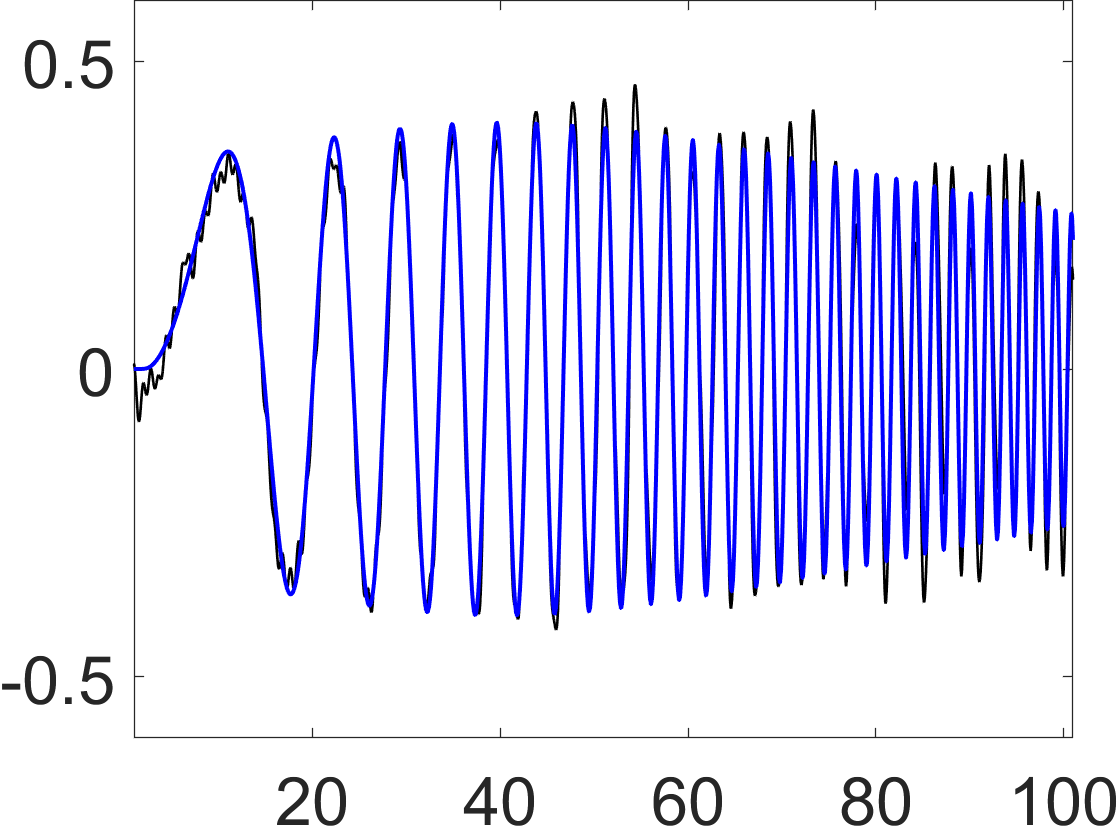}
  \caption{Fitting: chirp $\dot{q}_y$, constant rests}
  \label{subfig:dc_vy_only}
\end{subfigure}
\begin{subfigure}{0.115\linewidth}
  \centering
  \includegraphics[width=\linewidth]{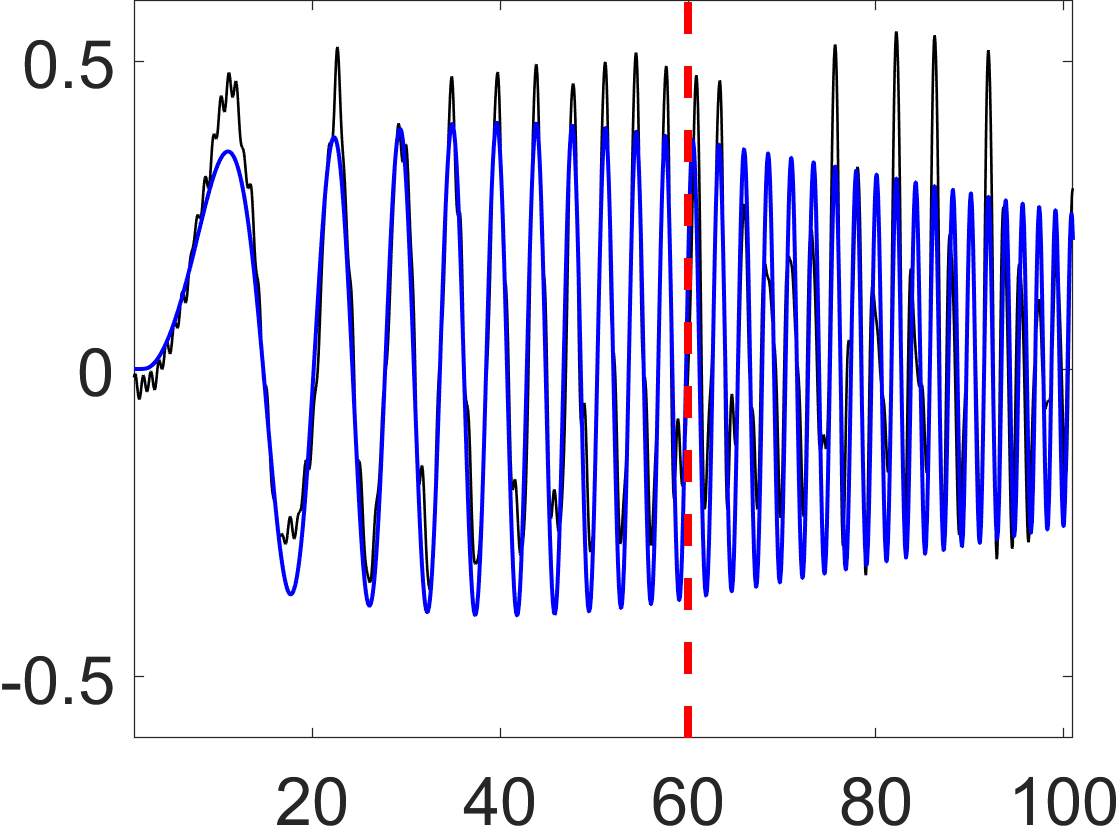}
  \caption{Testing input: chirp $\dot{q}_y$, chirp $\dot{q}_x$}
  \label{subfig:dc_vy_vxchirp}
\end{subfigure}
\begin{subfigure}{0.115\linewidth}
  \centering
  \includegraphics[width=\linewidth]{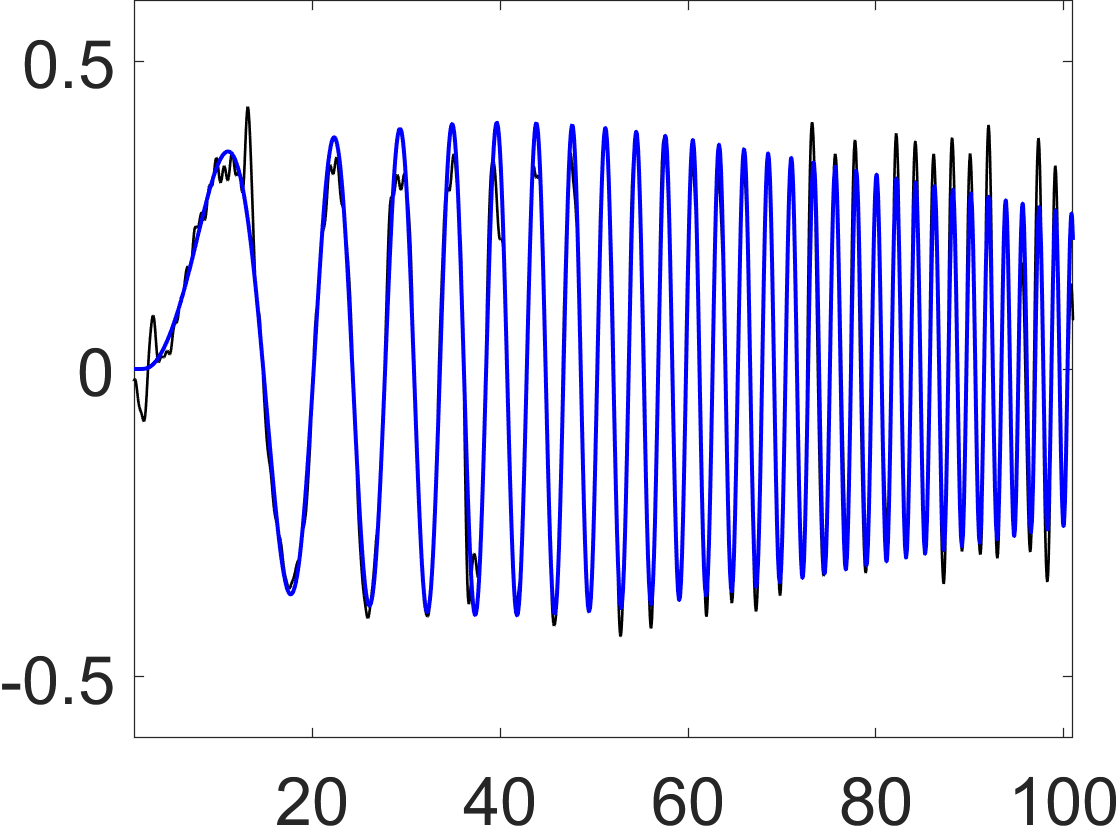}
  \caption{Testing input: chirp $\dot{q}_y$, chirp ${q}_z$}
  \label{subfig:dc_vy_whchirp}
\end{subfigure}
\begin{subfigure}{0.115\linewidth}
  \centering
  \includegraphics[width=\linewidth]{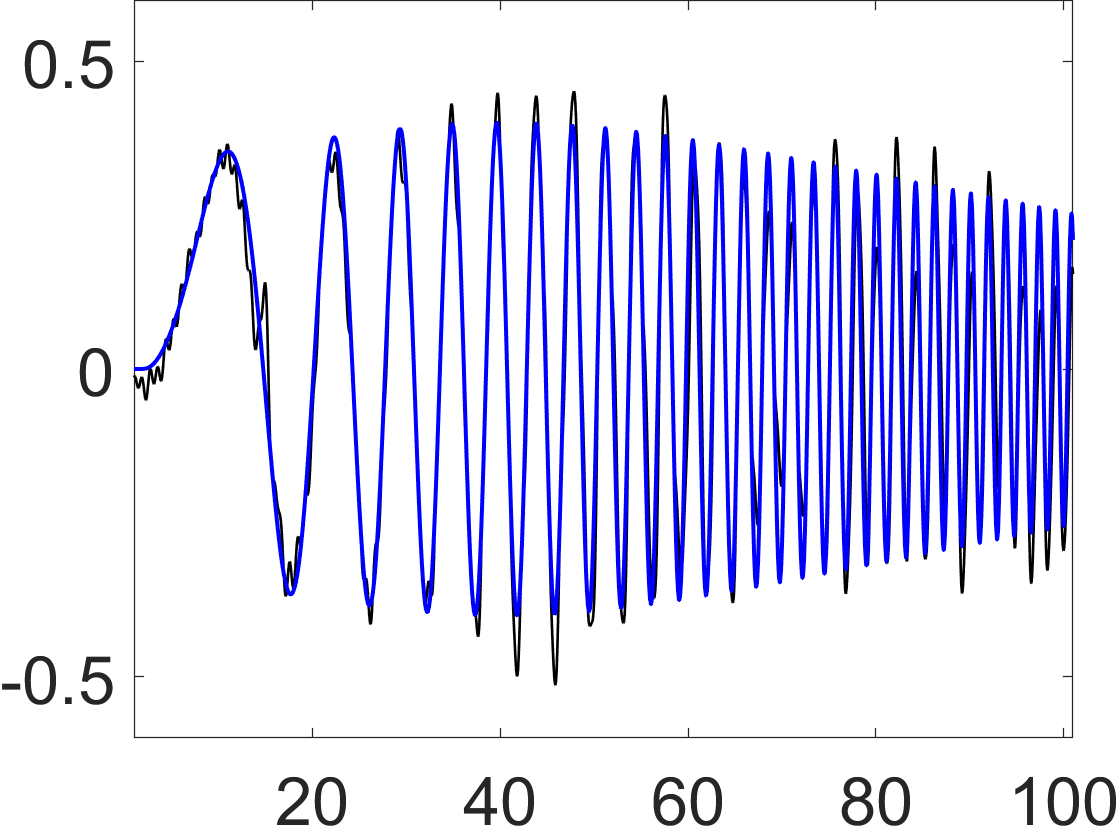}
  \caption{Testing input: chirp $\dot{q}_y$, chirp $\dot{q}_\phi$}
  \label{subfig:dc_vy_vyawchirp}
\end{subfigure}
\begin{subfigure}{0.115\linewidth}
  \centering
  \includegraphics[width=\linewidth]{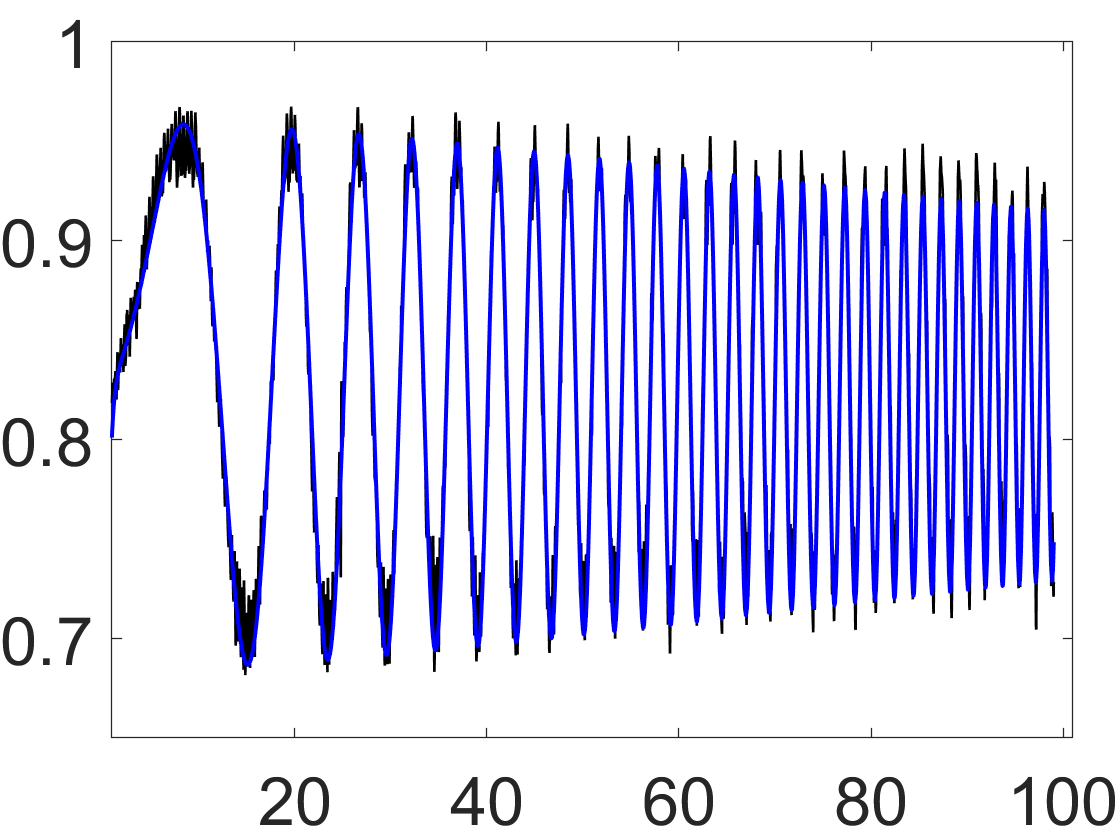}
  \caption{Fitting: chirp ${q}_z$, constant rests}
  \label{subfig:dc_wh_only}
\end{subfigure}
\begin{subfigure}{0.115\linewidth}
  \centering
  \includegraphics[width=\linewidth]{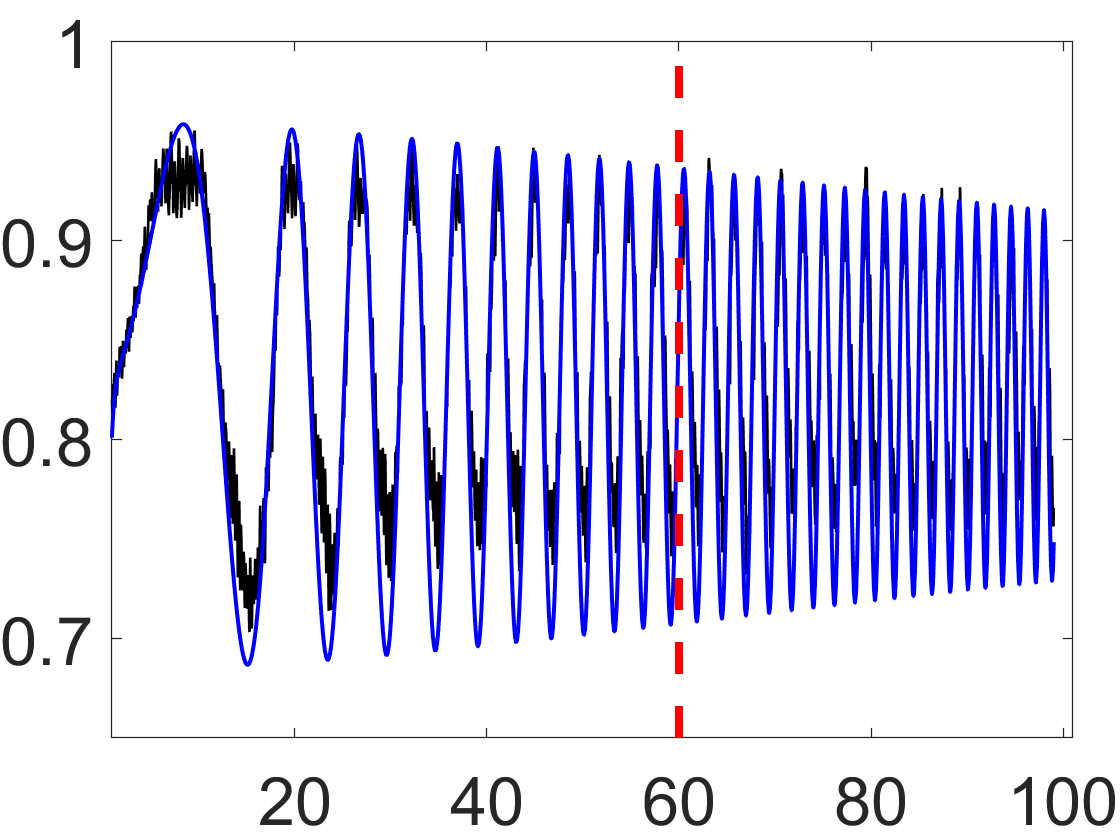}
  \caption{Testing input: chirp ${q}_z$, chirp $\dot{q}_x$}
  \label{subfig:dc_wh_vxchirp}
\end{subfigure}
\begin{subfigure}{0.115\linewidth}
  \centering
  \includegraphics[width=\linewidth]{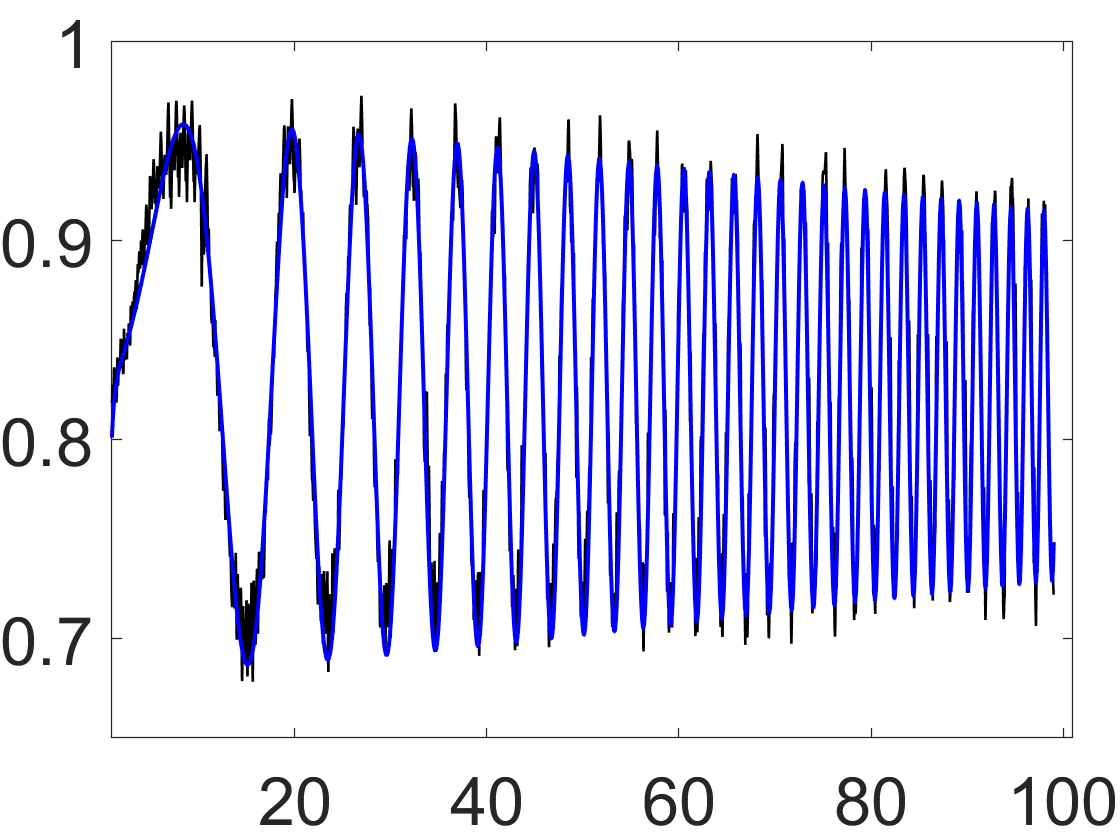}
  \caption{Testing input: chirp ${q}_z$, chirp $\dot{q}_y$}
  \label{subfig:dc_wh_vychirp}
\end{subfigure}
\begin{subfigure}{0.115\linewidth}
  \centering
  \includegraphics[width=\linewidth]{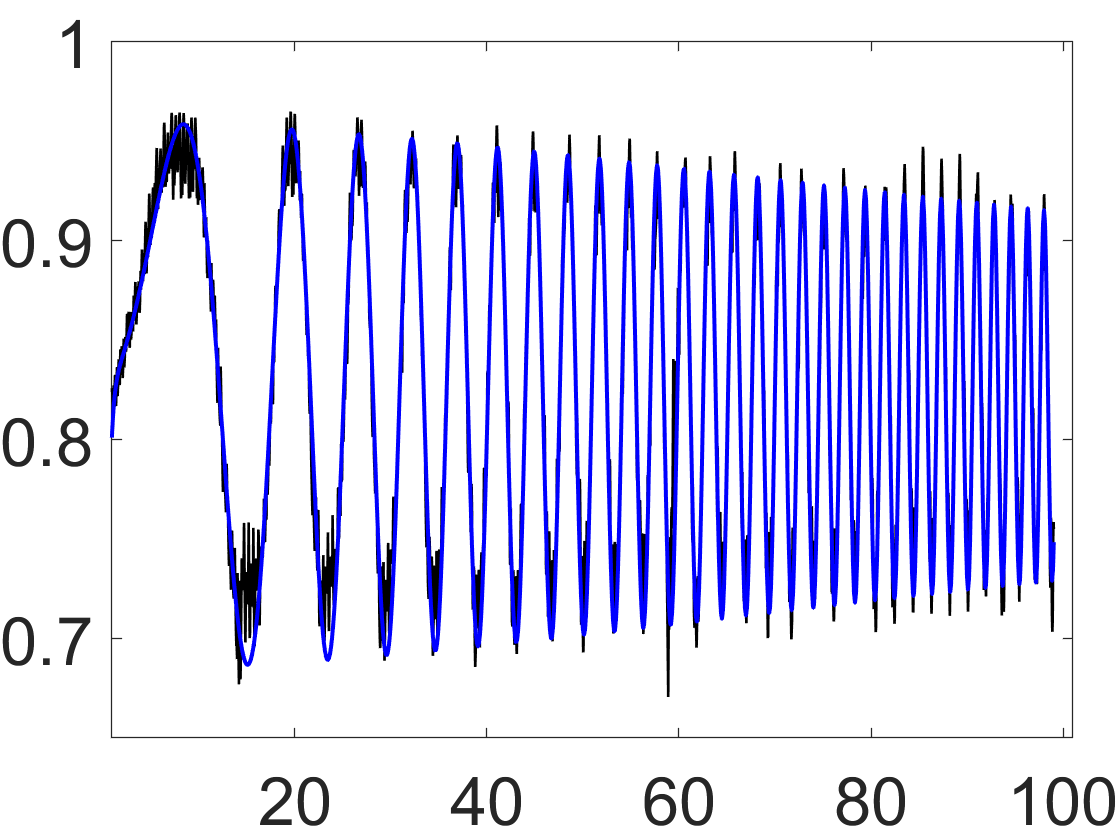}
  \caption{Testing input: chirp ${q}_z$, chirp $\dot{q}_\phi$}
  \label{subfig:dc_wh_vyawchirp}
\end{subfigure}
\begin{subfigure}{0.115\linewidth}
  \centering
  \includegraphics[width=\linewidth]{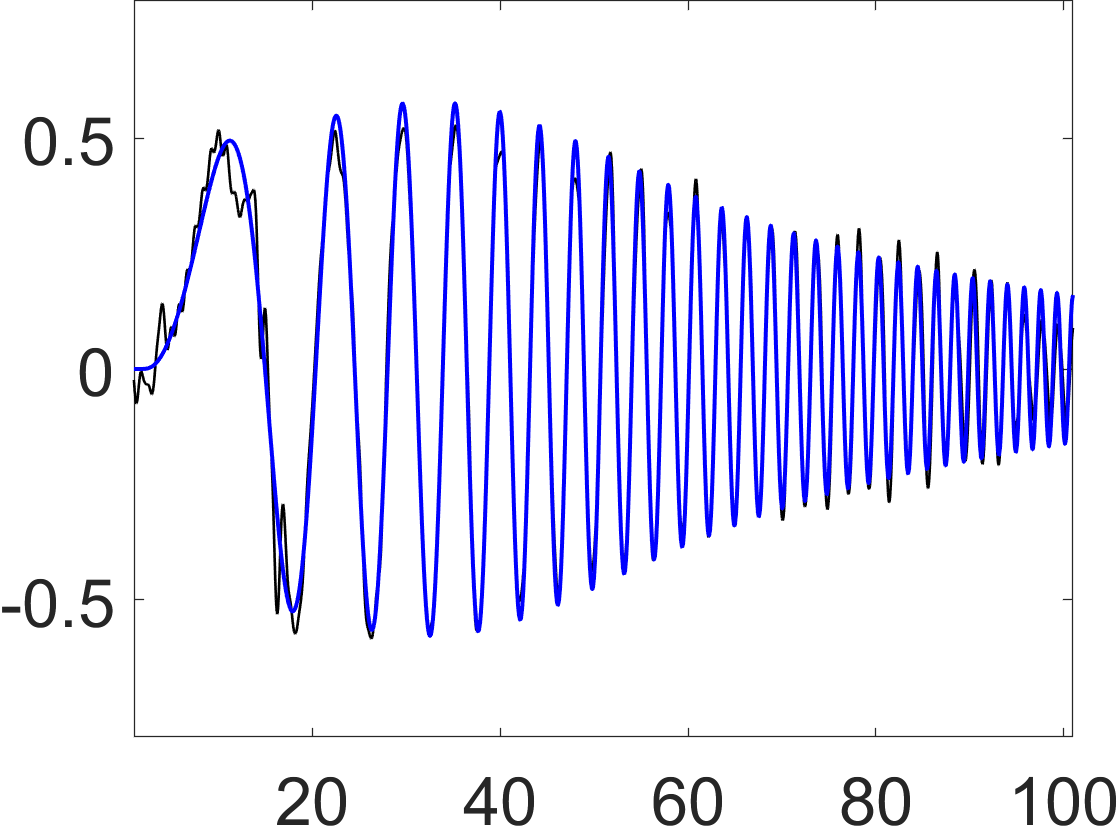}
  \caption{Fitting: chirp $\dot{q}_\phi$, constant rests}
  \label{subfig:dc_vyaw_only}
\end{subfigure}
\begin{subfigure}{0.115\linewidth}
  \centering
  \includegraphics[width=\linewidth]{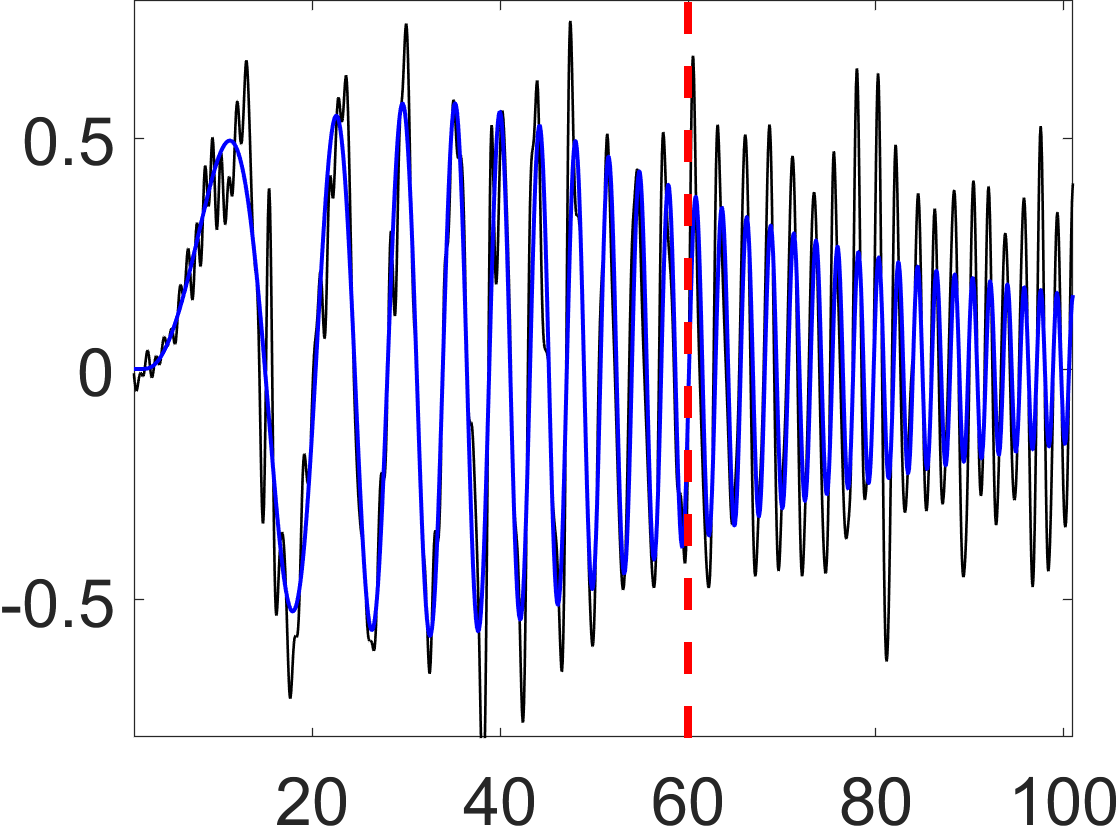}
  \caption{Testing input: chirp $\dot{q}_\phi$, chirp $\dot{q}_x$}
  \label{subfig:dc_vyaw_vxchirp}
\end{subfigure}
\begin{subfigure}{0.115\linewidth}
  \centering
  \includegraphics[width=\linewidth]{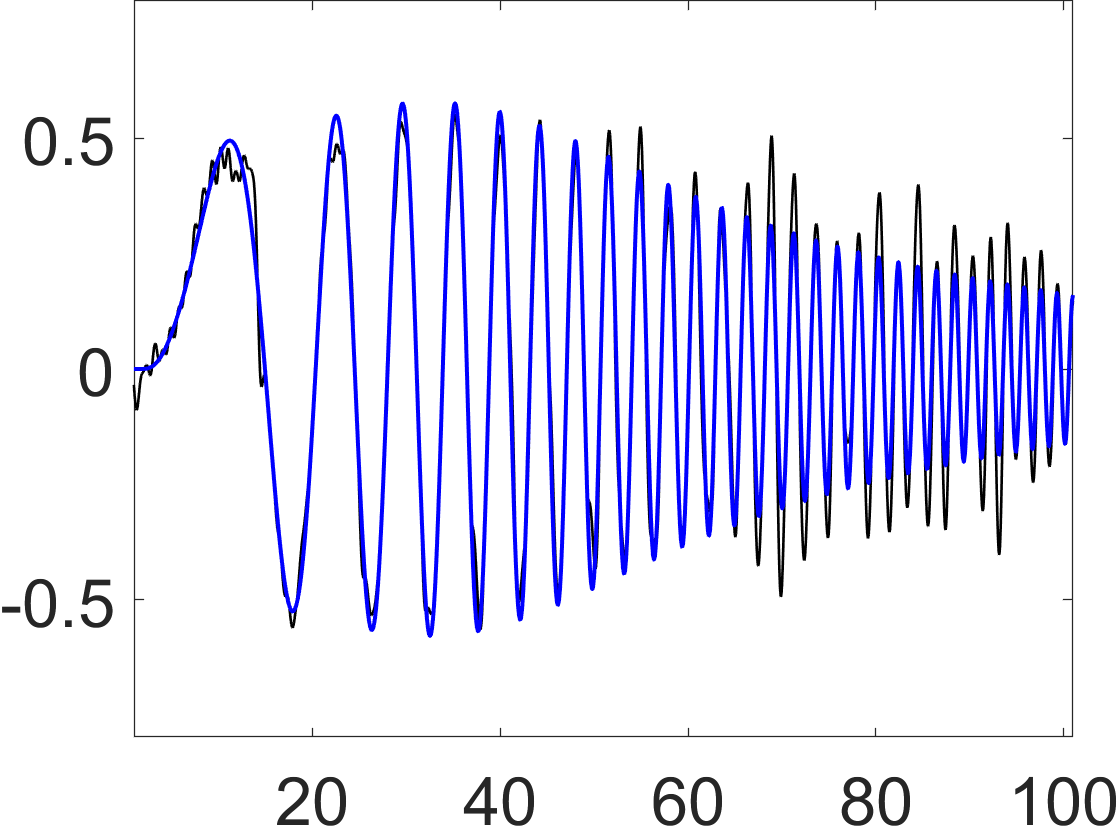}
  \caption{Testing input: chirp $\dot{q}_\phi$, chirp $\dot{q}_y$}
  \label{subfig:dc_vyaw_vychirp}
\end{subfigure}
\begin{subfigure}{0.115\linewidth}
  \centering
  \includegraphics[width=\linewidth]{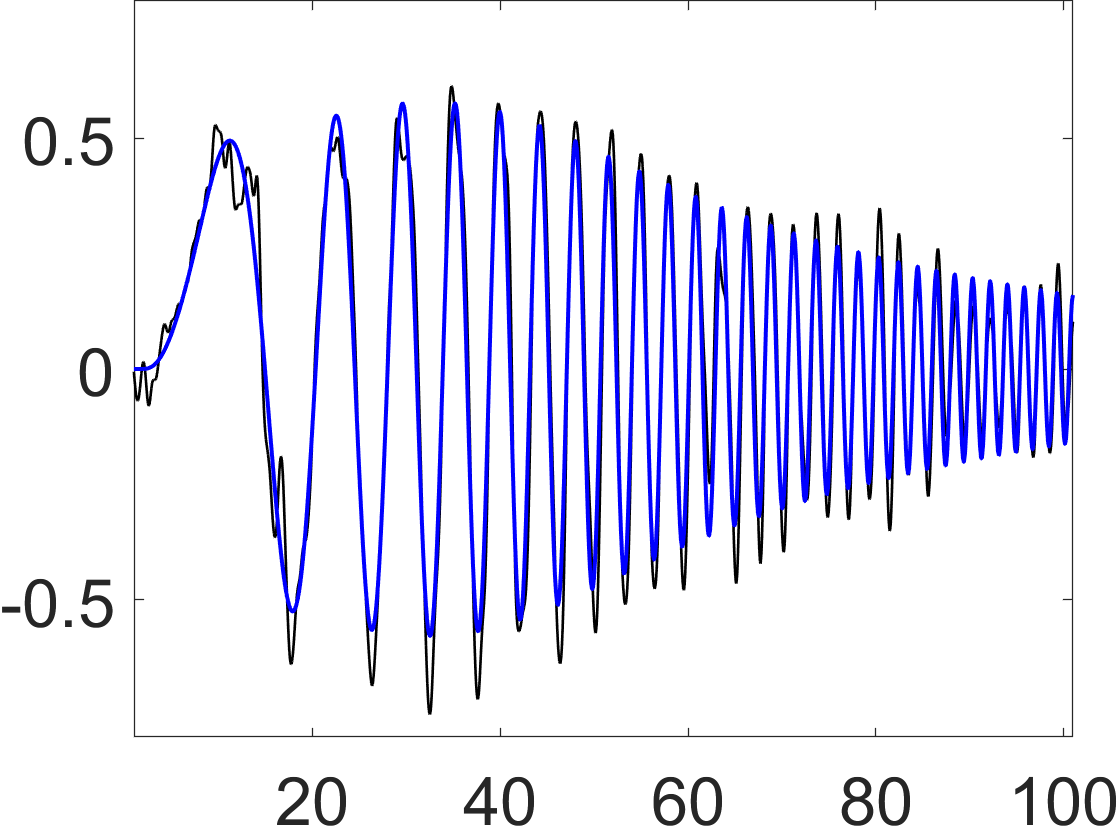}
  \caption{Testing input: chirp $\dot{q}_\phi$, chirp ${q}_z$}
  \label{subfig:dc_vyaw_whchirp}
\end{subfigure}
\caption{Decoupling test. The input to the system is a swept-frequency signal (chirp). The blue line represents for the predicted output excited by the input using the identified models in Sec.~\ref{sec:linearity} while black lines stands for the system's actual output. The red dash line shows the cutoff frequency $f_c = 0.6$~Hz. (a) The linear model found for $\dot{q}_x$ while other dimensions are constant,\textit{e.g.}, input for $\dot{q}_x$ is a chirp while the ones for $\dot{q}_y$, $\dot{q}_\phi$ are zeros and $q_z$ is 0.98~m (nominal height). (b)(c)(d) the same model is used to predict another measured input-output pair where the desired $\dot{q}_x$ and one other input dimension is also a chirp. $\dot{q}_x$ is independent of others since the identified model can well predict the measured outputs when other inputs are chirps. Same procedure is applied to other dimension pairs to show all the dimensions are decoupled such as (e)(f)(g)(h) to test the dependency of $\dot{q}_y$ to other dimensions, (i)(j)(k)(l) for ${q}_z$, (m)(n)(o)(p) for $\dot{q}_\phi$. Quantitative results are recorded in Tab.~\ref{tab:decouple_test_cnn}}
\label{fig:decoupling}
\end{figure*}

\begin{table*}[t]
\centering
\caption{Benchmark of Fitting Accuracy using the Identified Models on Different Input-Output Pairs: The header of each row indicates which dimension has a chirp input while others are constants during the model fitting. The column header stands for additional chirp input other than the row header to collect the input-output pairs to test the prediction accuracy of the system fitted for the row header. In this way, the diagonal of this table represents the prediction accuracy during the fitting stage while the rest is the accuracy during the testing stage.}
\label{tab:decouple_test_cnn}
\begin{tabular}{|c|c|c|c|c|}

\hline
 & $\mathbf{\dot{q}_x}$ \textbf{chirp} & $\mathbf{\dot{q}_y}$ \textbf{chirp} & $\mathbf{{q}_z}$ \textbf{chirp} & $\mathbf{\dot{q}_\phi}$ \textbf{chirp} \\ \hline
$\mathbf{\dot{q}_x}$ \textbf{chirp}    & $\mathbf{82.86\%}$, Fig.~\ref{subfig:dc_vx_only}   & $78.81\%$, Fig.~\ref{subfig:dc_vx_vychirp}        & $66.17\%$, Fig.~\ref{subfig:dc_vx_whchirp}         & $57.98\% (67.88\%^*)$, Fig.~\ref{subfig:dc_vx_vyawchirp}\\ \hline
$\mathbf{\dot{q}_y}$ \textbf{chirp}    & $45.65\% (69.36\%^*)$, Fig.~\ref{subfig:dc_vy_vxchirp} & $\mathbf{80.08\%}$, Fig.~\ref{subfig:dc_vy_only} & $80.65\%$, Fig.~\ref{subfig:dc_vy_whchirp}          & $77.76\%$, Fig.~\ref{subfig:dc_vy_vyawchirp}           \\ \hline
$\mathbf{{q}_z}$ \textbf{chirp}          & $47.85\% (54.29\%^*)$, Fig.~\ref{subfig:dc_wh_vxchirp}  & $81.12\%$, Fig.~\ref{subfig:dc_wh_vychirp}        & $\mathbf{83.19\%}$, Fig.~\ref{subfig:dc_wh_only} & $80.11\%$, Fig.~\ref{subfig:dc_wh_vyawchirp}         \\ \hline
$\mathbf{\dot{q}_\phi}$ \textbf{chirp} & $53.87\% (61.75\%^*)$, Fig.~\ref{subfig:dc_vyaw_vxchirp} & $79.39\%$, Fig.~\ref{subfig:dc_vyaw_vxchirp}        & $74.5\%$, Fig.~\ref{subfig:dc_vyaw_whchirp}          & $\mathbf{83.6\%}$, Fig.~\ref{subfig:dc_vyaw_only}   \\ \hline
\end{tabular}\\
*Indicates accuracy of the fit using the data before the proposed cut-off frequency $f_c=0.6$~Hz.
\end{table*}

The system, Cassie driven by the RL policy as a walking controller, is a Multiple-Input Multiple-Output~(MIMO) system, and there are four inputs, $\mathbf{u} \in \mathbb{R}^4$, and four outputs, $\mathbf{y} \in \mathbb{R}^4$. 
As demonstrated in an existing design for navigation autonomy on Cassie proposed in~\cite{li2021vision}, these four dimensions are tightly coupled using a model-based controller on Cassie and a lot of effort is exerted to consider such couplings for the planning purpose otherwise the planned output may cause a walking failure~\cite{li2021vision}.
In this section, we will show that four dimensional dynamics obtained in Sec.\ref{sec:linearity}, \textit{i.e.}, $f_{\dot{x}}(\mathbf{x}_{\dot{x}},{u}_{\dot{x}})$, $f_{\dot{y}}(\mathbf{x}_{\dot{y}},{u}_{\dot{y}})$, $f_{z}(\mathbf{x}_{z},{u}_{z})$, and $f_{\dot{\phi}}(\mathbf{x}_{\dot{\phi}},{u}_{\dot{\phi}})$, are all decoupled with respect to different inputs.

In order to test a system \textit{M} that is independent of another system \textit{N}, we applied four steps in two stages (fitting and testing stages):
\begin{enumerate}
    \item Collect the first measured input-output pair where the input to \textit{M} is a swept-frequency chirp signal, while the input to \textit{N} is a fixed constant value.
    \item Identify a model for \textit{M} using the first input-output pair. These two steps are in the fitting stage. 
    \item Obtain the second input-output pair where the input to \textit{M} is a chirp while the input to \textit{N} is also a chirp. 
    \item Test the prediction accuracy using the identified model from Step 2 on the second input-output pair from Step 3. Step 3 and 4 are in the testing stage.
\end{enumerate}
If the testing accuracy shows no significant loss than fitting accuracy, \textit{M} is independent of \textit{N}, otherwise, \textit{M} and \textit{N} are coupled.  
This is because a fixed value (zero frequency) doesn't contain any information of a swept-frequency wave (chirp) signal. 
As a result, if the model for \textit{M} identified by the data where the input to \textit{N} is a fixed value is able to accurately predict a measured input-output data when the input to \textit{N} is a chirp, this could show that the change in \textit{N} will not excite the dynamics in \textit{M}. 

\begin{figure*}[!]
\centering
\begin{subfigure}{0.245\linewidth}
  \centering
  \includegraphics[width=\linewidth]{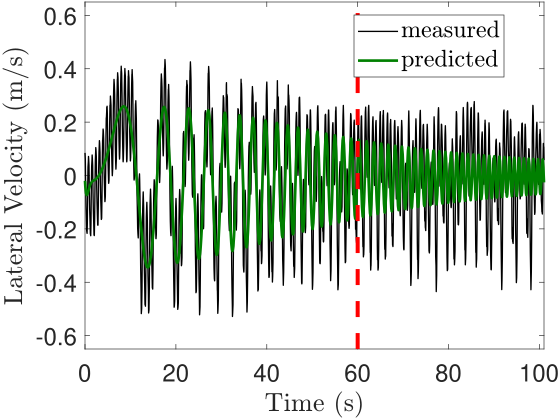}
  \caption{Fitted Model using Policy 1}
  \label{subfig:vy-md1}
\end{subfigure}
\begin{subfigure}{0.245\linewidth}
  \centering
  \includegraphics[width=\linewidth]{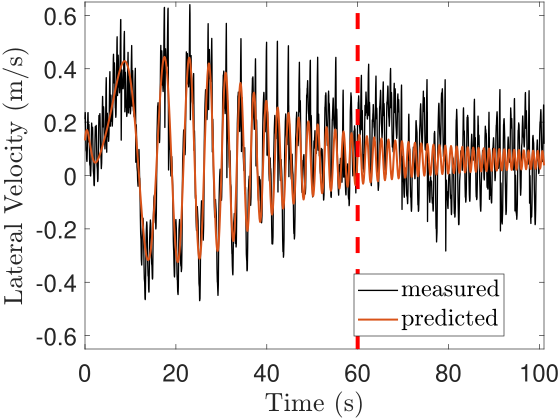}
  \caption{Fitted Model using Policy 2}
  \label{subfig:vy-md2}
\end{subfigure}
\begin{subfigure}{0.245\linewidth}
  \centering
  \includegraphics[width=\linewidth]{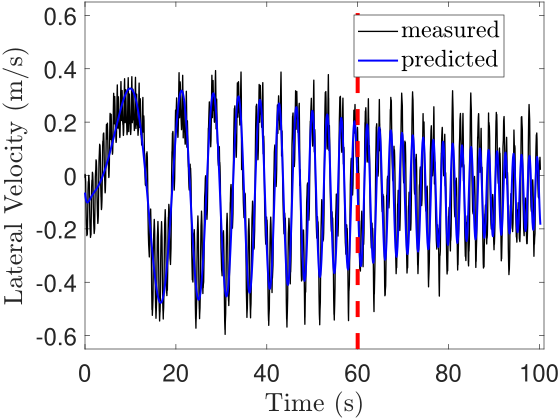}
  \caption{Fitted Model using Policy 3}
  \label{subfig:vy-md3}
\end{subfigure}
\begin{subfigure}{0.245\linewidth}
  \centering
  \includegraphics[width=\linewidth]{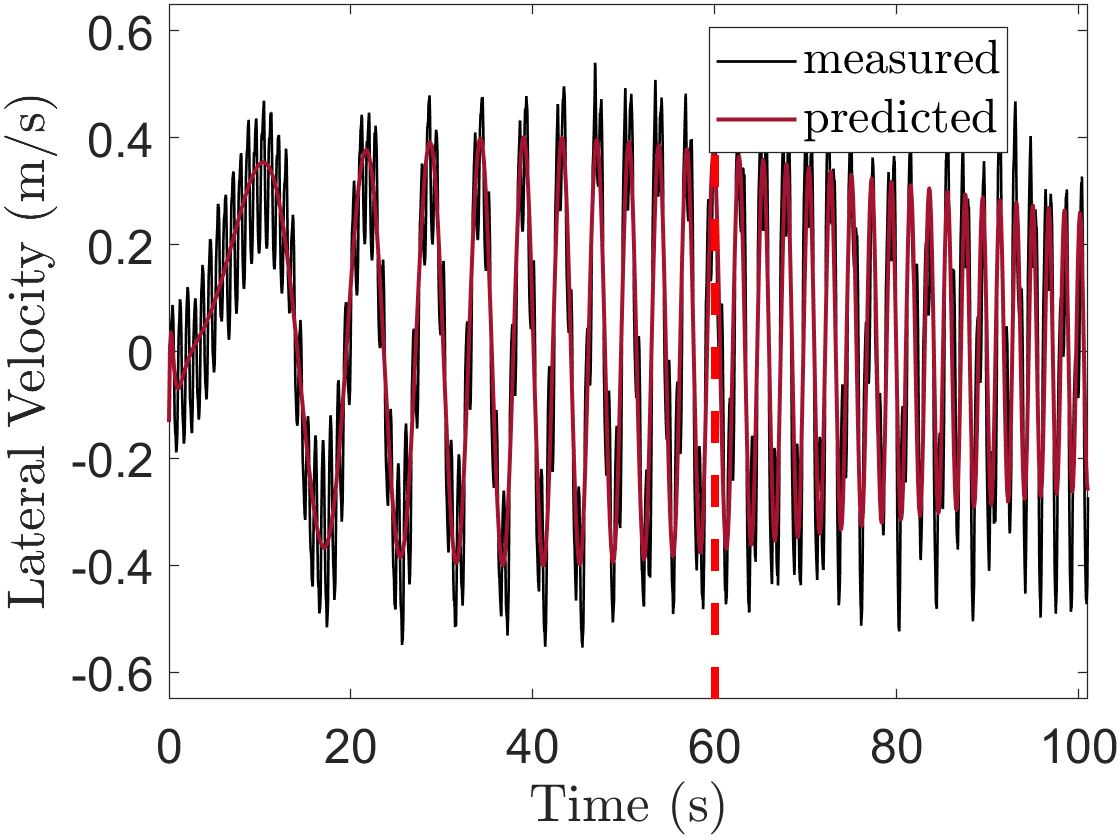}
  \caption{Fitted Model using Policy 4}
  \label{subfig:vy-md4}
\end{subfigure}
\caption{Cross policy validation using lateral walking velocity $\dot{q}_y$ as an example. Except the Policy 2 which is not well trained, although Cassie being controlled by different policies tends to show different extents of linearity, \textit{i.e.}, whether exist a linear model can fit the measured input-output well, all the policies enable the Cassie to demonstrate linearity before the frequency exceeding the cutoff frequency $f_c$ marked as red line.}
\label{fig:multi-model}
\end{figure*}

Let us consider the test on the dependency between sagittal walking velocity $\dot{q}_x$ and lateral walking velocity $\dot{q}_y$ using the CNN-based RL policy as an example.
During fitting the model for $\dot{q}_x$ in Sec.~\ref{sec:linearity}, the walking velocity command $\dot{q}_y$ is fixed at $0$~m/s, as shown in Fig.~\ref{subfig:dc_vx_only}.
The fitting accuracy is $82.86\%$. 
This fitted model is directly applied to predict the measured output of $\dot{q}_x$ when the input to $\dot{q}_y$ is also a chirp.
As shown in Fig.~\ref{subfig:dc_vx_vychirp}, the identified model still shows reasonably good prediction accuracy on this input-output pair, resulting in the accuracy of $78.81\%$. 
Therefore, we can draw a conclusion that sagittal walking velocity $\dot{q}_x$ is independent of lateral walking velocity $\dot{q}_y$.
Same decoupling tests are applied on all of the combinations of the four dimensions using the CNN-based RL policy, as shown in Fig.~\ref{fig:decoupling}, and the quantitative accuracy data is recorded in Tab.~\ref{tab:decouple_test_cnn}. 

According to Fig.~\ref{fig:decoupling} and Tab.~\ref{tab:decouple_test_cnn}, for the dynamics on each dimension, the model fitted on the data when the inputs on other dimensions are fixed values can well describe the dynamics when the input to one other dimension is a chirp, such as $\dot{q}_y \leftrightarrow q_z$ pair, $q_z \leftrightarrow \dot{q}_\phi$ pair, and $\dot{q}_y \leftrightarrow \dot{q}_\phi$ pair.
Although fitting accuracy for the $\dot{q}_y$, ${q}_z$ and $\dot{q}_\phi$ decreases under the existence of a chirp $\dot{q}_x$, the accuracy can be still over $50\%$ (the value with $^*$ in Tab.~\ref{tab:decouple_test_cnn}) if we only look at the measured output before the cut-off frequency $f_c=0.6$~Hz introduced in Sec.~\ref{subsec:criterion}. 
In conclusion, such a system can be considered as decoupled under a low-frequency input, and higher frequency may excite the coupled dynamics along some dimensions.  
\label{sec:decouple}

\section{Cross Policy Validation}

In order to show the generality of the phenomenon that when Cassie being controlled by a RL walking policy tends to show linearity, we test the measured input-output pairs obtained from using different RL policies.
We test $4$ policies: 
\begin{enumerate}
    \item A MLP-based RL policy is well trained with around $400$ million samples from scratch using reward formulation.
    \item A MLP-based RL policy uses a different reward form but is in the middle of training after $30$ million samples using Policy $1$ as a warm start.
    \item A MLP-based RL Policy keeps training on Policy $2$ using the same reward and is trained with $100$ million samples.
    \item A CNN-based RL policy (different neural network structure) and different reward, and well trained after $400$~M samples from scratch.
\end{enumerate} 
We use the input-output dynamics of lateral walking velocity ($\hat{\dot{q}}_y = f_{\dot{y}}(\mathbf{x}_{\dot{y}},u_{\dot{y}})$) as an example.  
The system identification is applied on Cassie driven by these 4 policies, respectively, and the results are shown in Fig.~\ref{fig:multi-model}.
If the measured input-output data can be well fitted by a linear model, we can tell that Cassie being controlled by each of these policies shows linearity. 

According to Fig.~\ref{fig:multi-model}, since Policy $2$ is not well trained, the system driven by this policy shows the worst linearity, \textit{i.e.}, a linear model cannot well predict the system output by the input.
Specifically, the nonlinearity appears after the cut-off frequency $f_c$. 
After being trained with ample samples, Policy 3 which uses the same reward and neural network structure, shows a clear existence of the linearity in Fig.~\ref{subfig:vy-md3}. 
Moreover, the measured input-output pairs from the other policies, like Policy $1$ and $4$, demonstrate reasonably good fitting accuracy before cutoff frequency $f_c=0.6$~Hz.

From all the discussion above, we can draw a conclusion that linearity is not always guaranteed. 
If a RL policy is not well-trained, such as Policy 2, even if we “force" the system identification to find a linear structure, the fitting result could be poor because of the existence of strong nonlinearity in the system. 
However, if the learning of the policies has converged, the linearity appears, such as Fig.~\ref{subfig:vy-md1},~\ref{subfig:vy-md3},~\ref{subfig:vy-md4}, and this is validated across different RL policies with different rewards or neural network structure. 
\begin{remark}
The rewards used in this paper do not include a term to encourage the robot to behave like a linear system.
But encoding a desired linear system behavior in the reward could be an interesting future work. 
\end{remark}

\begin{remark}
We also note that the proposed linearity analysis could potentially be a metric for the convergence of the learning of a RL policy on a dynamic system.
\end{remark}

Therefore, we can summarize the low-dimensional representation of Cassie being controlled by a well-trained RL policy using a linear model as below:
\noindent
\begin{equation}
\label{eq:linear-model}
\begin{bmatrix}
    \mathbf{\dot{x}} \\ 
    \mathbf{y} \\
\end{bmatrix}=  
\begin{bmatrix}
    A & B \\ 
    C & 0 \\
\end{bmatrix}
\begin{bmatrix}
    \mathbf{x} \\ 
    \mathbf{u} \\ 
\end{bmatrix}
\end{equation}
\noindent
where the states $\mathbf{x} = [\mathbf{x}_{\dot{x}}^T, \mathbf{x}_{\dot{y}}^T, \mathbf{x}_{z}^T, \mathbf{x}_{\dot{\theta}}^T]^T$, output $\mathbf{y}$ is $[y_{\dot{x}}, y_{\dot{y}}, y_{z}, y_{\dot{\theta}}]^T = [\dot{q}_x, \dot{q}_y, q_z, \dot{q}_\theta]^T$, and input $\mathbf{u}$ is $ [u_{\dot{x}}, u_{\dot{y}}, u_{z}, u_{\dot{\theta}}]^T = [\dot{q}^d_x, \dot{q}^d_y, q^d_z, \dot{q}^d_\theta]^T$, and matrices $A=\text{diag}(A_{\dot{x}}, A_{\dot{y}}, A_{z}, A_{\dot{\theta}})$, $B$ and $C$ are proper stack of $B_{\dot{x}}, B_{\dot{y}}, B_{z}, B_{\dot{\theta}}$ and $C_{\dot{x}}, C_{\dot{y}}, C_{z}, C_{\dot{\theta}}$, respectively. Hence the linearized dynamics is as follows,
\begin{equation}
\begin{bmatrix}
    \mathbf{\dot{x}}_{\dot{x}, \dot{y}, z, \dot{\phi}} \\ 
    \mathbf{y}_{\dot{x}, \dot{y}, z, \dot{\phi}} \\
\end{bmatrix}=  
\begin{bmatrix}
    A_{\dot{x}, \dot{y}, z, \dot{\phi}} & B_{\dot{x}, \dot{y}, z, \dot{\phi}} \\ 
    C_{\dot{x}, \dot{y}, z, \dot{\phi}} & 0 \\
\end{bmatrix}
\begin{bmatrix}
    \mathbf{x}_{\dot{x}, \dot{y}, z, \dot{\phi}} \\ 
    \mathbf{u}_{\dot{x}, \dot{y}, z, \dot{\phi}} \\ 
\end{bmatrix}.
\end{equation}
\noindent
These models are the control canonical forms of \eqref{eq:vx_dynamics}, \eqref{eq:vy_dynamics}, \eqref{eq:wh_dynamics}, and \eqref{eq:vyaw_dynamics} shown in Sec.~\ref{sec:linearity}, respectively. 
\label{sec:multiple}

\section{Case Study: Safe Navigation for Bipedal Robots in Height-Constrained Environment}

\subsection{Safety-critical Navigation Framework}
In this section, we demonstrate an application of the proposed methodology, that is, utilizing the identified simple system to provide safety guarantees on the nonlinear system driven by its RL policy.

\begin{figure}
    \centering
    \includegraphics[width=\linewidth]{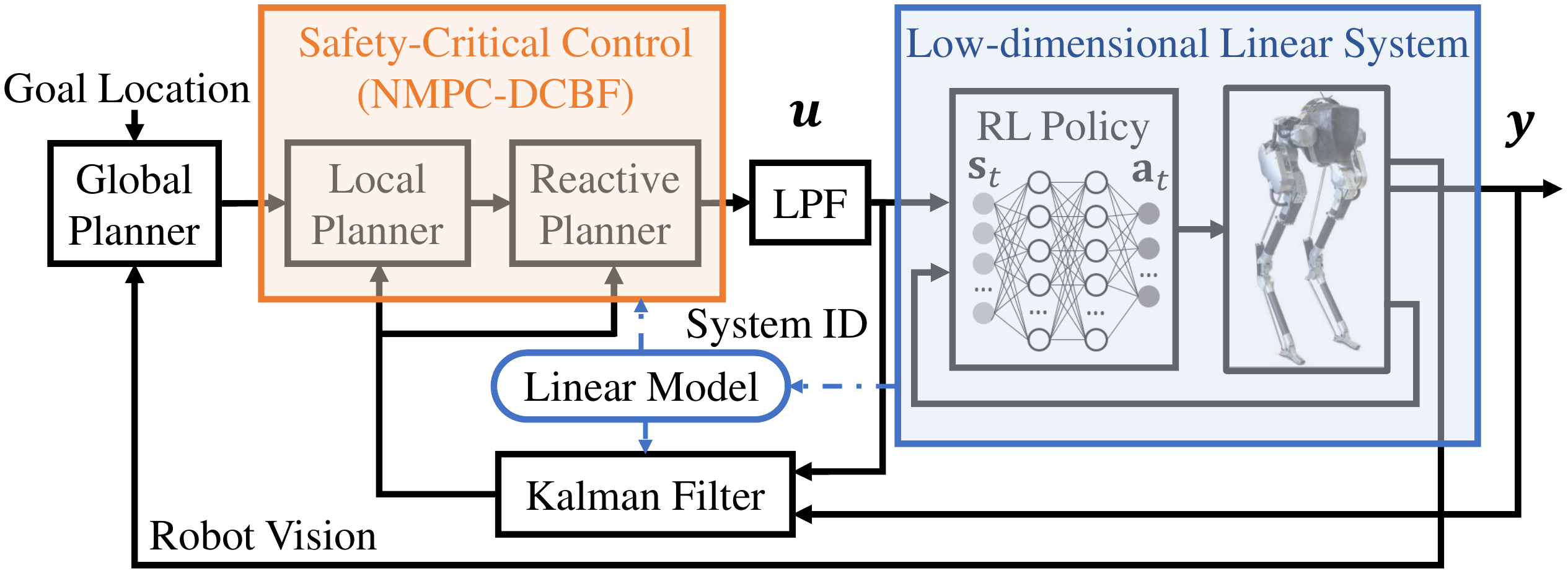}
    \caption{Proposed safe navigation autonomy using Cassie driven by a RL walking policy and the identified linear system in this work. The linear model of Cassie being controlled by RL is identified and used by the local planner and the reactive planner. Those planners are formulated by using a safety-critical control framework NMPC-DCBF. The output of the reactive planner is passed through a Low Pass Filter~(LPF) with cut off frequency $0.5$~Hz in order to prevent violating the linearity criterion of the identified system. The states of the linear system are estimated by a Kalman Filter.}
    \label{fig:framework}
\end{figure}

The linear system representation expressed in \eqref{eq:linear-model} can be applied to solve a safe navigation problem for Cassie. 
The navigation autonomy framework proposed in this paper can be found in Fig.~\ref{fig:framework} which is based on~\cite{li2021vision} to navigate unknown environments with height-constraints using Cassie.
In this autonomy, after being given a goal location, a global planner firstly finds a collision-free path on the map from the robot's current position, followed by a local planner and a reactive planner using nonlinear optimization to track the global path.
The locomotion controller used in the previous work to enable Cassie to follow planning results is based on Hybrid Zero Dynamics~(HZD) and is developed in~\cite{li2020animated} based on~\cite{gong2019feedback}.
A detailed introduction of this autonomy can be found in~\cite{li2021vision} which serves as a baseline in this work.

In this work, we use the CNN-based RL locomotion policy on Cassie. 
Instead of using the nonlinear reduced-order dynamics model in~\cite{li2021vision} during constrained local planning by collocation, we adopt its identified linear system in~\eqref{eq:linear-model} and safety-critical control framework that combines nonlinear model predictive control~(NMPC) and discrete-time control barrier function (DCBF) introduced in~\cite{zeng2021enhancing}. 
In this way, we can combine the advantages of the model-free RL which can bring us robust and agile controller on a complex nonlinear system~\cite{li2021reinforcement} while using its identified linear system is more practical to utilize online in the NMPC-DCBF framework to provide enhanced feasibility and safety performance~\cite{zeng2021enhancing}. 

To formulate the NMPC-DCBF problem, we over-approximate the geometries of Cassie and the obstacles by cylinders.
The distance between robot and each obstacle $o_i$ can be computed analytically as follows,
\begin{equation}
d^{o_i}_k = (x^g_k - x^{o_i})^2 + (y^g_k - y^{o_i})^2 - (R^{robot} + R^{o_i})^2,
\label{eq:dcbf-cassie-obstacles}
\end{equation}
where $x^g_k$ $y^g_k$ are the global x-y position of the robot at time $k$, $(x^{o_i}, y^{o_i})$ is the position of the obstacle $o_i$, $R^{robot}$ is the radius of the cylindrical approximation of Cassie, and $R^{o_i}$ is the radius of the obstacle.
Obstacle avoidance between the robot and the obstacle can then be enforced by constraining $d^{o_i}_k > 0, \forall k$.
However, over short planning horizons this constraint can fail to ensure long-term obstacle avoidance.
Therefore, we use DCBF constraints, which can provide long-term obstacle avoidance, even on short planning horizons~\cite{zeng2021safety}.
The DCBF constraint is as follows,
\begin{equation}
    d^{o_i}_{k+1} \geq \omega_k \alpha_{\text{DCBF}} d^{o_i}_{k},   0 \leq \alpha_{\text{DCBF}} \leq 1, \omega_k \geq 0,
\end{equation}
where $\alpha_{\text{DCBF}}$ represents the maximum decay-rate at which $d^{o_i}_k$ can converge to zero and $\omega_k$ is a slack variable which enhances feasibility and safety~\cite{zeng2021enhancing}.
Specifically, the trajectory generation problem with NMPC-DCBF is formulated as follows,
\begin{subequations}
\begin{align}
        \min_{\mathbf{x}, \mathbf{u}, \bm{\delta}} ~& J (\mathbf{x}, \mathbf{u}, \bm{\delta}, \bm{\omega}),  \label{eq:problem} \\
    \mbox{s.t.}~~&\mathbf{x}_{k+1} = A^d \mathbf{x}_k + B^d\mathbf{u}_k \label{eq:discrete-time-dynamics}\\
    &\mathbf{x}_{0} = \mathbf{x}_{init}\label{eq:init-constraint} \\
    &\mathbf{x}_k \in \mathcal{X}_{adm}, \mathbf{u}_k \in \mathcal{U}_{adm}, \label{eq:state-input-constraint}\\
    & x^{g}_{k+1} = x^{g}_{k} + C^d_{\dot{x}}\mathbf{x}_{\dot{x}, k} \cos(C^d_{\dot{\phi}}\mathbf{x}_{\phi, k}) dt, \label{eq:kinematics-constraint-x}\\
    & y^{g}_{k+1} = y^{g}_{k} + C^d_{\dot{y}}\mathbf{x}_{\dot{y}, k} \sin(C^d_{\dot{\phi}}\mathbf{x}_{\phi, k}) dt, \label{eq:kinematics-constraint-y} \\
    & d^{o_i}_{k+1} \geq \omega_k \alpha_{\text{DCBF}} d^{o_i}_{k}, \quad \omega_k \geq 0, \label{eq:dcbf-constraint} \\
    & [x^{g}_{N}, y^{g}_{N}, \mathbf{x}_{\phi, N}]^T = [x^{g}_{f}, y^{g}_{f}, \mathbf{x}_{\phi, f}]^T +  \bm{\delta}
\end{align}
\end{subequations}
where $A^d$, $B^d$, $C^d$ are coefficients of discrete-time dynamics transferred from the proposed continuous linear dynamics in \eqref{eq:linear-model}. The cost function is designed as below,
\begin{equation}
\begin{split}
    J (\mathbf{x}, \mathbf{u}, \bm{\delta}, \bm{\omega}) 
    = &\sum_{i=1}^{N-1} (||\mathbf{x}_i||^2_{Q} + ||\mathbf{u}_i||^2_{R} + ||\mathbf{x}_{i+1} - \mathbf{x}_i||^2_{dQ} \\ & + \rho (1-\omega_k)^2) + ||\bm{\delta}||^{2}_{K}.
\end{split}
\label{eq:cost-function}
\end{equation}
The discrete-time dynamics \eqref{eq:discrete-time-dynamics}, 
together with kinematics along x and y axis in \eqref{eq:kinematics-constraint-x} and \eqref{eq:kinematics-constraint-x} formulate a trajectory optimization problem, where the robot's orientation is considered to transfer the robot's velocity in the robot's frame into the world frame.
The initial condition, state and input constraint can be found in \eqref{eq:init-constraint}, \eqref{eq:state-input-constraint}.
Specifically, $\mathcal{X}_{adm}$ represents the combination of state constraint together with safety constraint to avoid nearby obstacles defined by distance functions, presented in \cite{li2021vision}.
$\mathcal{U}_{adm}$ is the input constraint.
$\bm{\omega} = [\omega_1, ..., \omega_{N-1}]^T$ represents the relaxation variables of decay rate $\alpha_{\text{DCBF}}$ for DCBF constraints. We also notice that $\rho$ in \eqref{eq:cost-function} shall be chosen as a relatively large scalar such that the DCBF constraints wouldn't be over-relaxed~\cite{zeng2021pointwise}.
Moreover, $\bm{\delta}$ represents the slack variable for terminal constraint on desired final state $[x^{g}_{f}, y^{g}_{f}, \mathbf{x}_{\phi, f}]^T$ which is minimized with a quadratic term in the cost function.

\begin{figure*}
    \centering
    \begin{subfigure}{0.27\textwidth}
        \centering
        \includegraphics[width=\linewidth]{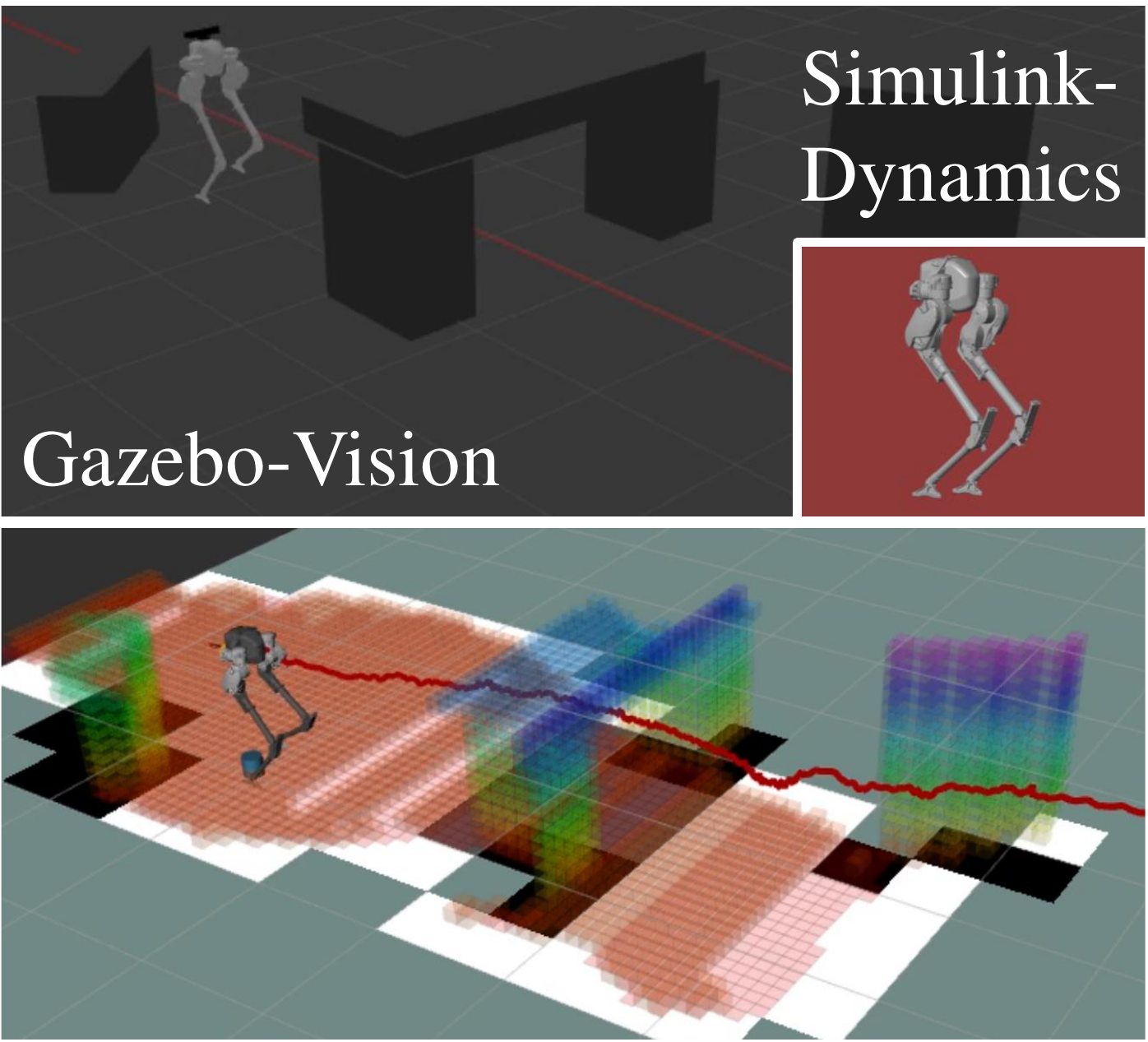}
        \caption{Joint simulation of vision and dynamics using the safe navigation autonomy in a congested space} \label{subfig:sim_vis}
    \end{subfigure} 
    \begin{subfigure}{0.275\textwidth}
        \centering
        \includegraphics[width=\linewidth]{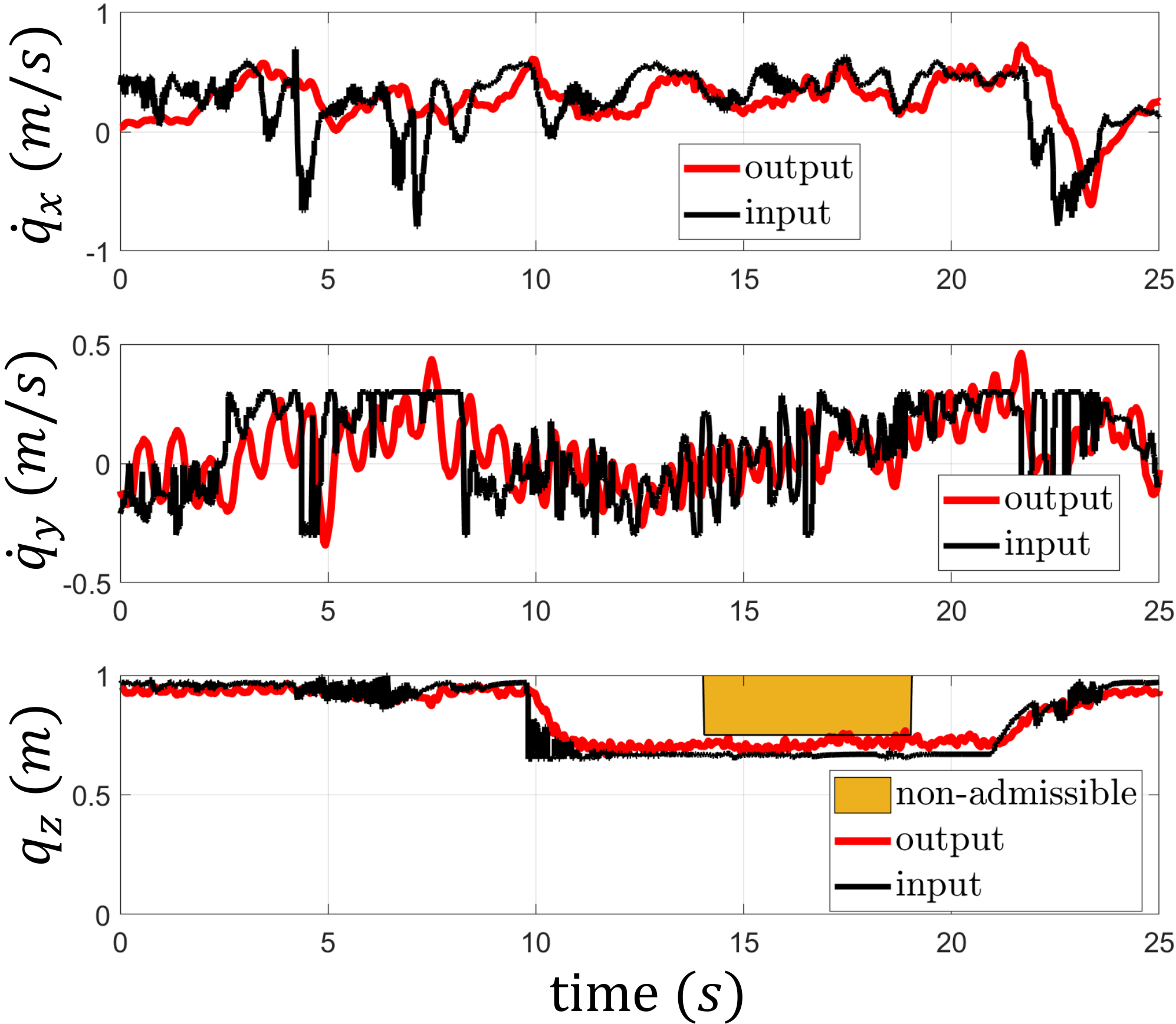}
        \caption{The input (desired commands) and output (robot measured states) during navigation in Fig.~\ref{subfig:sim_vis}} \label{subfig:sim_plot}
    \end{subfigure}
    \begin{subfigure}{0.375\textwidth}
        \centering
        \includegraphics[width=\linewidth]{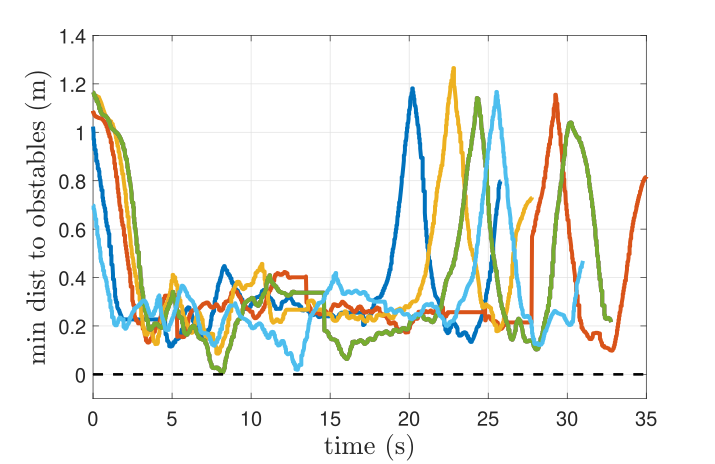}
        \caption{The recorded distance between the robot and its closet obstacles for $5$ repeated tests in the same scenario but with randomized initial conditions} \label{subfig:safety_analysis}
    \end{subfigure}   
    \caption{Validation of the proposed safety-critical navigation autonomy in the joint simulation. (a) The robot's depth camera reading is simulated in Gazebo while robot's dynamics is synchronously computed in Simulink. The planning data is also visualized, such as the robot travelled path marked as red line. (b) The proposed safe navigation framework enables Cassie using its agile RL locomotion controller to quickly arrive to the goal location and crouch down to travel underneath an arch without a single collision. (c) The tests are repeated $5$ times with randomized initial conditions. Through all of these tests, the robot-obstacle separation never goes below $0$ m after subtracting the shape of the robot and obstacles. Video can be found at \url{https://sites.google.com/berkeley.edu/rl-sysid-rss2022}.}    
    \label{fig:result}
\end{figure*}

The output from the local planner is a two-second dynamic-feasible and collision-free trajectory, while the output from the reactive planner is the real-time desired sagittal and lateral walking velocities $\dot{q}^d_{x,y}$, walking height $q^d_z$, and turning yaw rate $\dot{q}^d_\phi$.
This is the input $\mathbf{u}$ to the linear system that was identified to describe the closed-loop dynamics of Cassie driven by its RL locomotion policy.
Moreover, $\mathbf{u}$ is firstly passed through a low pass filter with cutoff frequency of $0.5$~Hz which is below the threshold of the linearity criterion found in Sec.~\ref{subsec:criterion} to prevent exciting nonlinearity of the system.
\begin{remark}
Since there is a modeling mismatch between the ground truth model (the closed loop system) and the identified linear system, we need to make the CBF robust to modelling errors. However, the robust form of MPC with DCBF with horizons is still an unaddressed problem~\cite{cheng2020safe}.
Therefore, in this paper, we include a safety buffer to the obstacles. 
We mark developing robust MPC-DCBF as an important future work to provide a formal safety guarantee on the closed system controlled by RL.
\end{remark}

\subsection{Autonomy Validation}


To validate the proposed autonomy that combines NMPC-DCBF with CNN-based RL locomotion policy, the entire algorithm is tested in a joint simulation on Cassie, as shown in Fig.~\ref{subfig:sim_vis}
In this test, a congested space with two obstacles and a height-constrained space (arch) in between is built in Gazebo where the robot depth camera reading is simulated. 
The robot dynamics driven by its RL policy is computed in MATLAB Simulink which has high-fidelity for the dynamics computation.
These two simulators are synchronized and the result is illustrated in Fig.~\ref{fig:result}.

As demonstrated in Fig.~\ref{subfig:sim_vis},\ref{subfig:sim_plot}, during the simulation, the robot successfully achieves accelerating to full speed (around $1$~m/s) when there are no obstacles while quickly pulling back when obstacles are in range.
Moreover, the robot shows the capacity to crouch down to travel underneath the arch with a relative high speed. 
By using the linear model in the optimization, the two-second local trajectory can be solved in around $0.1 - 0.2$~s which is at least $5$~times faster than the prior approach using a nonlinear model~\cite{li2021vision}.
Additionally, it turns out that there are reduced deadlocks and the whole navigation task can be finished in around $25$~s which is almost twice faster than the prior work with HZD-based controller~\cite{li2021vision} which results in a conservative planned speed (it finishes the same trial in $50$~s).
For the controller performance, the RL-based policy for walking controller is also more robust and agile than the HZD-based walking controller, such as the coupled walking dynamics are cancelled by RL so the robot can quickly lift its body up after passing through an arch while accelerating to its goal without falling over, as shown in Fig.~\ref{subfig:sim_vis}.
To demonstrate the provided safety, we run the test 5 times and record the distance between the robot and its closet obstacles in Fig.~\ref{subfig:safety_analysis}. 
During all of these tests in the same scenario with different robot initial conditions, the robot never collides with the obstacle as the distance between the robot and its closet obstacle is always above $0$~m after considering robot and obstacle shapes, which indicates the safety is preserved empirically. 

By using nonlinear model predictive control with control barrier functions, we provide a safety guarantee on such a high order nonlinear complex system.
All this allows Cassie controlled by RL locomotion policy during navigation while enabling the safety-critical control-planners to exploit the agility brought by the RL policy.

\label{sec:application}

\section{Conclusion and Future Works}
In conclusion, for the task of velocity and height tracking control of a bipedal robot, we have presented evidence that a model-free reinforcement learning based controller acting on a highly nonlinear dynamical system may tend to linearize it such that the entire closed-loop system can be represented by a low-dimensional linear system.
This is an interesting finding since a high order nonlinear system controlled by a high dimensional nonlinear neural network policy behaves like a linear system.
Based on this observation, we propose a new direction to bridge safety-critical control and model-free RL by finding certifications for stability and safety on the low-dimensional model identified on the complex system driven by its RL policy. 

We validate this methodology on a bipedal robot Cassie controlled by its RL locomotion policy.
We apply system identification on this closed-loop system to find a linear model, and later we find that such a system demonstrates reasonably good linearity. 
Moreover, the fitted multiple-input multiple-output linear model is decoupled, minimum phase and asymptotically stable in all dimensions.
We also provide a criterion for the linearity existence, that is, the input frequency should be under a given threshold. 
By cross comparing such linearity on different RL policies, we show that linearity is preserved across different well-trained policies, but linearity is not guaranteed if the learning of the RL policy has not converged.

The proposed linearity analysis on the nonlinear system controlled by RL policy can serve two purposes: control and learning. 
For the control purpose, finding and analyzing the linearity property of the system driven by RL can help us to understand its limitation and provide guidance to design the high-level controller. 
This can make a RL policy be utilized for higher level of autonomy.
For the learning purpose, we show that the linearity can be a metric for learning convergence, as we see in this work, linearity may not appear if the policy is not well trained. 

For application, the fitted linear model is later utilized in a safety-critical navigation framework using NMPC-DCBF which utilizes the RL policy as a low-level walking controller on Cassie.
In this application, we note that while the RL policy for the walking controller is able to address the highly nonlinear dynamics of Cassie, the identified closed-loop dynamics represented in a linear model is able to provide guarantees of stability while also introducing safety through control barrier functions online.

However, we also note that the proposed method may not be general to all nonlinear systems and model-free RL algorithms/tasks. 
It may be only applicable to the systems that are feedback linearizable. 
Future work could extend such linearity analysis on other nonlinear systems and exploit the usability of a low-dimensional linear system on other existing model-based safety-critical control and planning methods such as HJ reachability~\cite{chen2021fastrack}, dynamic programming~\cite{bertsekas2011dynamic} to provide safety and stability guarantee for Cassie in a variety of tasks in the real world.
Moreover, since most of analysis made in this paper is numerical, mathematical analysis with proofs about existence of low-dimensional representations on general nonlinear dynamical systems with RL approaches would also be a fundamental contribution to the community.
\label{sec:conclusion}

\section*{Acknowledgements}
This work was supported in part by NSF Grant CMMI-1944722. The authors would like to thank Prof. Shankar Sastry, Prof. Yi Ma, Prof. Claire Tomlin, Prof. Glen Berseth, and Prof. Xue Bin Peng for providing insightful discussions and comments on this work. The authors also want to thank all of the anonymous reviewers for their critical feedback.

\appendix{
\subsection{Fitting Results for MLP-based RL Policy}
\label{subsec:fitting-results-mlp}

We present the fitting results using linear models for all four dimensions of Cassie being controlled by the MLP controller in Figure \ref{fig:id_result_mlp}.

\begin{figure}
\centering
\begin{subfigure}{0.49\linewidth}
  \centering
  \includegraphics[width=\linewidth]{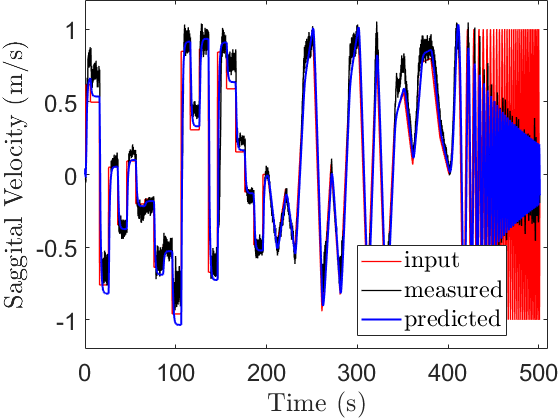}
  \caption{Fitted Linear Model of $\dot{q}_x$}
  \label{subfig:vx_id_mlp}
\end{subfigure}
\begin{subfigure}{0.49\linewidth}
  \centering
  \includegraphics[width=\linewidth]{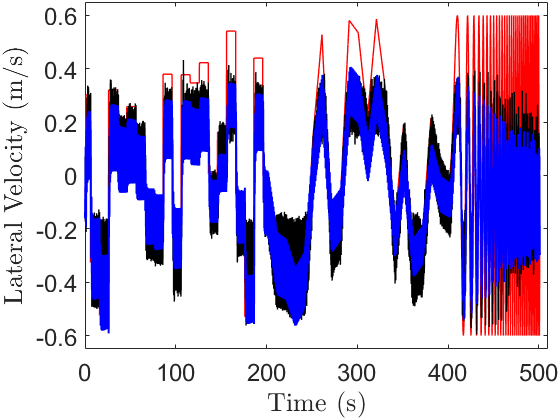}
  \caption{Fitted Linear Model of $\dot{q}_y$}
  \label{subfig:vy_id_mlp}
\end{subfigure}
\begin{subfigure}{0.49\linewidth}
  \centering
  \includegraphics[width=\linewidth]{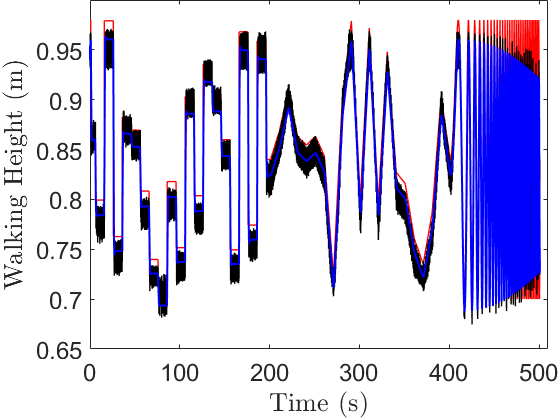}
  \caption{Fitted Linear Model of $q_z$}
  \label{subfig:wh_id_mlp}
\end{subfigure}
\begin{subfigure}{0.49\linewidth}
  \centering
  \includegraphics[width=\linewidth]{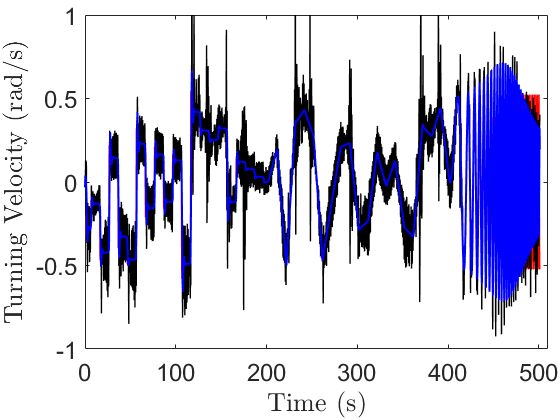}
  \caption{Fitted Linear Model of $\dot{q}_\phi$}
  \label{subfig:vyaw_id_mlp}
\end{subfigure}
\caption{Fitting results using linear models for all four dimension of Cassie being controlled by the MLP controller. The fitting accuracy for saggital walking velocity, lateral walking velocity, walking height, and turning yaw rate are $78.8\%$, $64.22\%$, $86.5\%$, and $59.03\%$, respectively. }
\label{fig:id_result_mlp}
\end{figure}

\subsection{Pole-Zero Plot For Identified Linear Systems of MLP-based RL Policy}

We present the pole-zero plot for the identified linear systems for all four dimensions of Cassie being controlled by the MLP controller in Figure \ref{fig:zero-pole-mlp}.

\label{subsec:pole-zero-plot-mlp}
\begin{figure}
\centering
\begin{subfigure}{0.32\linewidth}
  \centering
  \includegraphics[width=\linewidth]{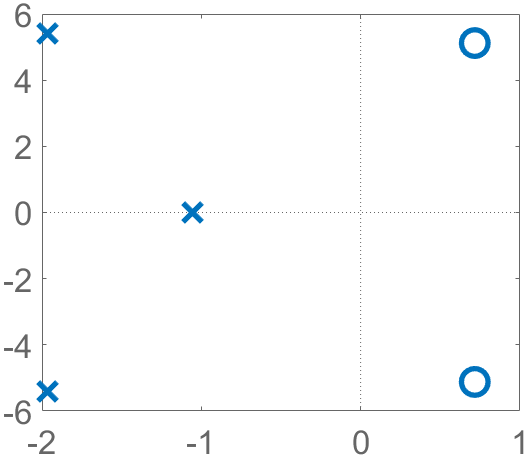}
  \caption{Pole-Zero Plot of $\dot{q}_x$}
  \label{subfig:vx_zp_mlp}
\end{subfigure}
\begin{subfigure}{0.32\linewidth}
  \centering
  \includegraphics[width=\linewidth]{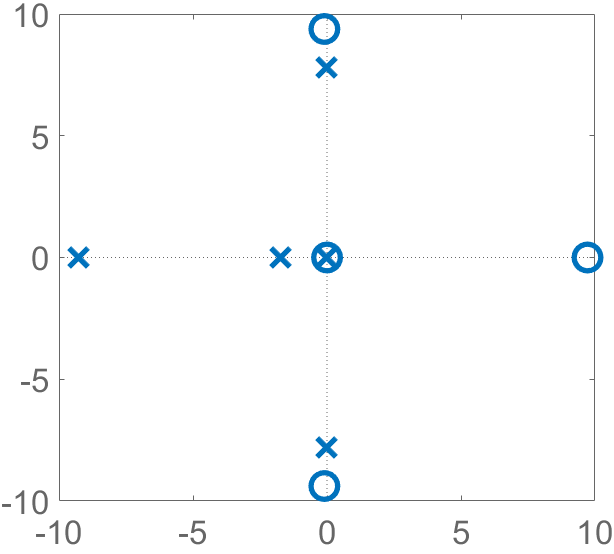}
  \caption{Pole-Zero Plot of $\dot{q}_y$}
  \label{subfig:vy_zp_mlp}
\end{subfigure}
\begin{subfigure}{0.32\linewidth}
  \centering
  \includegraphics[width=\linewidth]{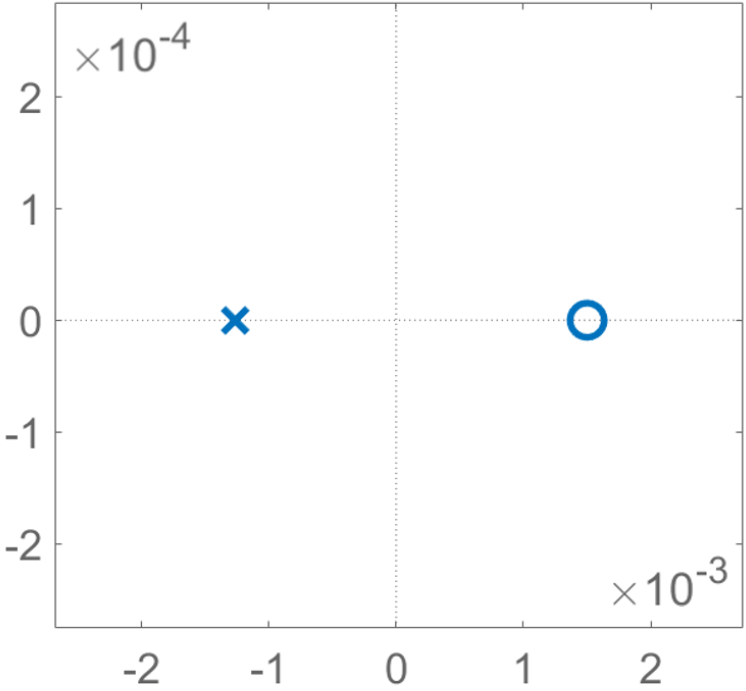}
  \caption{Zoomed Plot of $\dot{q}_y$}
  \label{subfig:vy_zp_zoom_mlp}
\end{subfigure}
\begin{subfigure}{0.32\linewidth}
  \centering
  \includegraphics[width=\linewidth]{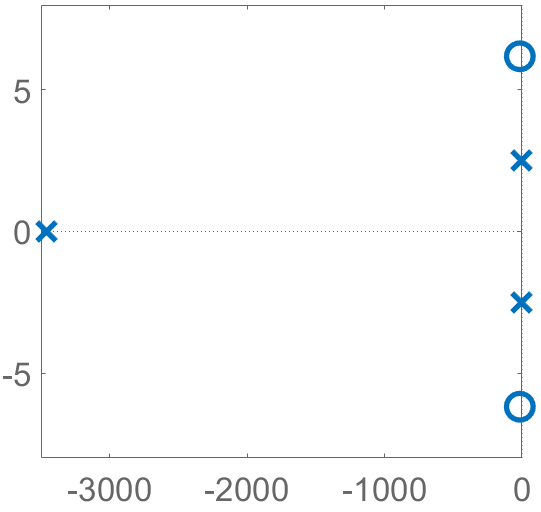}
  \caption{Pole-Zero Plot of $q_z$}
  \label{subfig:wh_zp_mlp}
\end{subfigure}
\begin{subfigure}{0.32\linewidth}
  \centering
  \includegraphics[width=\linewidth]{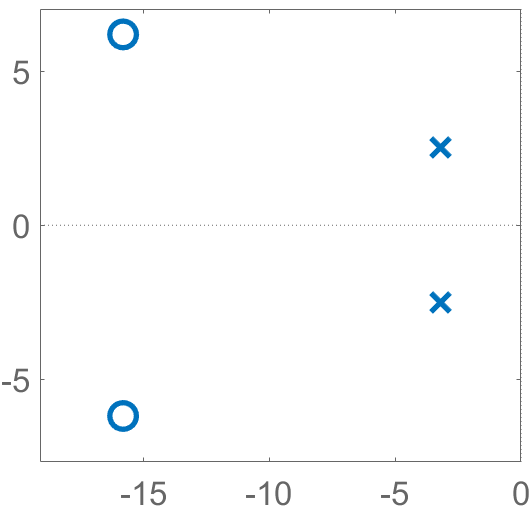}
  \caption{Zoomed Plot of $q_z$}
  \label{subfig:wh_zp_zoom_mlp}
\end{subfigure}
\begin{subfigure}{0.32\linewidth}
  \centering
  \includegraphics[width=\linewidth]{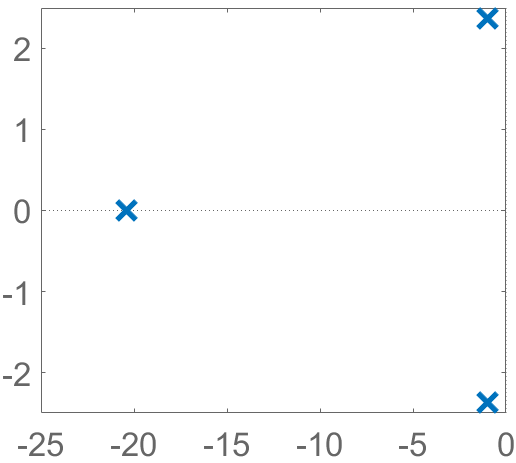}
  \caption{Pole-Zero Plot of $\dot{q}_\phi$}
  \label{subfig:vyaw_zp_mlp}
\end{subfigure}
\caption{Pole-Zero plots of the identified linear systems of MLP controller for saggital walking velocity, lateral walking velocity, walking height, and turning yaw rate, respectively. All poles are on the left hand plane~(LHP), but there are unstable zeros for the $\dot{q}_x$ dynamics and $\dot{q}_y$ dynamics, \textit{i.e.}, the system is non-minimum phase.}
\label{fig:zero-pole-mlp}
\end{figure}

}

\bibliographystyle{plainnat}
\bibliography{bib/bibliography}

\begin{thebibliography}{74}
\providecommand{\natexlab}[1]{#1}
\providecommand{\url}[1]{\texttt{#1}}
\expandafter\ifx\csname urlstyle\endcsname\relax
  \providecommand{\doi}[1]{doi: #1}\else
  \providecommand{\doi}{doi: \begingroup \urlstyle{rm}\Url}\fi

\bibitem[Allen-Zhu et~al.(2019)Allen-Zhu, Li, and Song]{allen2019convergence}
Zeyuan Allen-Zhu, Yuanzhi Li, and Zhao Song.
\newblock A convergence theory for deep learning via over-parameterization.
\newblock In \emph{International Conference on Machine Learning}, pages
  242--252. PMLR, 2019.

\bibitem[Alshiekh et~al.(2018)Alshiekh, Bloem, Ehlers, K{\"o}nighofer, Niekum,
  and Topcu]{alshiekh2018safe}
Mohammed Alshiekh, Roderick Bloem, R{\"u}diger Ehlers, Bettina K{\"o}nighofer,
  Scott Niekum, and Ufuk Topcu.
\newblock Safe reinforcement learning via shielding.
\newblock \emph{Proceedings of the AAAI Conference on Artificial Intelligence},
  32\penalty0 (1), 2018.

\bibitem[Ames et~al.(2014)Ames, Grizzle, and Tabuada]{ames2014control}
Aaron~D Ames, Jessy~W Grizzle, and Paulo Tabuada.
\newblock Control barrier function based quadratic programs with application to
  adaptive cruise control.
\newblock In \emph{53rd IEEE Conference on Decision and Control}, pages
  6271--6278, 2014.

\bibitem[Ames et~al.(2019)Ames, Coogan, Egerstedt, Notomista, Sreenath, and
  Tabuada]{ames2019control}
Aaron~D Ames, Samuel Coogan, Magnus Egerstedt, Gennaro Notomista, Koushil
  Sreenath, and Paulo Tabuada.
\newblock Control barrier functions: Theory and applications.
\newblock In \emph{2019 18th European Control Conference (ECC)}, pages
  3420--3431, 2019.

\bibitem[Bansal and Tomlin(2021)]{bansal2021deepreach}
Somil Bansal and Claire~J. Tomlin.
\newblock Deepreach: A deep learning approach to high-dimensional reachability.
\newblock In \emph{2021 IEEE International Conference on Robotics and
  Automation (ICRA)}, pages 1817--1824, 2021.

\bibitem[Bansal et~al.(2017)Bansal, Chen, Herbert, and
  Tomlin]{bansal2017hamilton}
Somil Bansal, Mo~Chen, Sylvia Herbert, and Claire~J Tomlin.
\newblock Hamilton-jacobi reachability: A brief overview and recent advances.
\newblock In \emph{2017 IEEE 56th Annual Conference on Decision and Control
  (CDC)}, pages 2242--2253, 2017.

\bibitem[Bertsekas(2011)]{bertsekas2011dynamic}
Dimitri~P Bertsekas.
\newblock Dynamic programming and optimal control, volume ii.
\newblock \emph{Belmont, MA: Athena Scientific}, 2011.

\bibitem[Boffi et~al.(2021)Boffi, Tu, Matni, Slotine, and
  Sindhwani]{boffi2020learning}
Nicholas Boffi, Stephen Tu, Nikolai Matni, Jean-Jacques Slotine, and Vikas
  Sindhwani.
\newblock Learning stability certificates from data.
\newblock In \emph{Proceedings of the 2020 Conference on Robot Learning},
  volume 155 of \emph{Proceedings of Machine Learning Research}, pages
  1341--1350. PMLR, 16--18 Nov 2021.

\bibitem[Bujarbaruah et~al.(2018)Bujarbaruah, Zhang, Rosolia, and
  Borrelli]{bujarbaruah2018adaptive}
Monimoy Bujarbaruah, Xiaojing Zhang, Ugo Rosolia, and Francesco Borrelli.
\newblock Adaptive mpc for iterative tasks.
\newblock In \emph{2018 IEEE Conference on Decision and Control (CDC)}, pages
  6322--6327, 2018.

\bibitem[Castillo et~al.(2021)Castillo, Weng, Zhang, and
  Hereid]{castillo2021robust}
Guillermo~A. Castillo, Bowen Weng, Wei Zhang, and Ayonga Hereid.
\newblock Robust feedback motion policy design using reinforcement learning on
  a 3d digit bipedal robot.
\newblock In \emph{2021 IEEE/RSJ International Conference on Intelligent Robots
  and Systems (IROS)}, pages 5136--5143, 2021.

\bibitem[Chan et~al.(2021)Chan, Yu, You, Qi, Wright, and Ma]{chan2021redunet}
Kwan Ho~Ryan Chan, Yaodong Yu, Chong You, Haozhi Qi, John Wright, and Yi~Ma.
\newblock Redunet: A white-box deep network from the principle of maximizing
  rate reduction.
\newblock \emph{arXiv preprint arXiv:2105.10446}, 2021.

\bibitem[Chen and Tomlin(2018)]{chen2018hamilton}
Mo~Chen and Claire~J Tomlin.
\newblock Hamilton--jacobi reachability: Some recent theoretical advances and
  applications in unmanned airspace management.
\newblock \emph{Annual Review of Control, Robotics, and Autonomous Systems},
  1:\penalty0 333--358, 2018.

\bibitem[Chen et~al.(2021)Chen, Herbert, Hu, Pu, Fisac, Bansal, Han, and
  Tomlin]{chen2021fastrack}
Mo~Chen, Sylvia Herbert, Haimin Hu, Ye~Pu, Jaime~Fernandez Fisac, Somil Bansal,
  SooJean Han, and Claire~J Tomlin.
\newblock Fastrack: a modular framework for real-time motion planning and
  guaranteed safe tracking.
\newblock \emph{IEEE Transactions on Automatic Control}, 2021.

\bibitem[Cheng et~al.(2019)Cheng, Orosz, Murray, and Burdick]{cheng2019end}
Richard Cheng, G{\'a}bor Orosz, Richard~M Murray, and Joel~W Burdick.
\newblock End-to-end safe reinforcement learning through barrier functions for
  safety-critical continuous control tasks.
\newblock \emph{Proceedings of the AAAI Conference on Artificial Intelligence},
  33\penalty0 (01):\penalty0 3387--3395, 2019.

\bibitem[Cheng et~al.(2020)Cheng, Khojasteh, Ames, and Burdick]{cheng2020safe}
Richard Cheng, Mohammad~Javad Khojasteh, Aaron~D Ames, and Joel~W Burdick.
\newblock Safe multi-agent interaction through robust control barrier functions
  with learned uncertainties.
\newblock In \emph{2020 59th IEEE Conference on Decision and Control (CDC)},
  pages 777--783, 2020.

\bibitem[Choi et~al.(2020)Choi, Castaneda, Tomlin, and
  Sreenath]{choi2020reinforcement}
Jason Choi, Fernando Castaneda, Claire~J Tomlin, and Koushil Sreenath.
\newblock Reinforcement learning for safety-critical control under model
  uncertainty, using control lyapunov functions and control barrier functions.
\newblock In \emph{Proceedings of Robotics: Science and Systems}, 2020.

\bibitem[Chou et~al.(2021)Chou, Ozay, and Berenson]{chou2021model}
Glen Chou, Necmiye Ozay, and Dmitry Berenson.
\newblock Model error propagation via learned contraction metrics for safe
  feedback motion planning of unknown systems.
\newblock \emph{arXiv preprint arXiv:2104.08695}, 2021.

\bibitem[Chow et~al.(2017)Chow, Ghavamzadeh, Janson, and Pavone]{chow2017risk}
Yinlam Chow, Mohammad Ghavamzadeh, Lucas Janson, and Marco Pavone.
\newblock Risk-constrained reinforcement learning with percentile risk
  criteria.
\newblock \emph{The Journal of Machine Learning Research}, 18\penalty0
  (1):\penalty0 6070--6120, 2017.

\bibitem[Chow et~al.(2018)Chow, Nachum, Duenez-Guzman, and
  Ghavamzadeh]{chow2018lyapunov}
Yinlam Chow, Ofir Nachum, Edgar Duenez-Guzman, and Mohammad Ghavamzadeh.
\newblock A lyapunov-based approach to safe reinforcement learning.
\newblock \emph{Advances in neural information processing systems}, 31, 2018.

\bibitem[Dai et~al.(2021)Dai, Landry, Yang, Pavone, and
  Tedrake]{dai2021lyapunov}
Hongkai Dai, Benoit Landry, Lujie Yang, Marco Pavone, and Russ Tedrake.
\newblock Lyapunov-stable neural-network control.
\newblock In \emph{Proceedings of Robotics: Science and Systems}, 2021.

\bibitem[Dawson et~al.(2022)Dawson, Qin, Gao, and Fan]{dawson2022safe}
Charles Dawson, Zengyi Qin, Sicun Gao, and Chuchu Fan.
\newblock Safe nonlinear control using robust neural lyapunov-barrier
  functions.
\newblock In \emph{Conference on Robot Learning}, pages 1724--1735. PMLR, 2022.

\bibitem[Fan et~al.(2020)Fan, Agha-mohammadi, and Theodorou]{fan2020deep}
David~D Fan, Ali-akbar Agha-mohammadi, and Evangelos~A Theodorou.
\newblock Deep learning tubes for tube mpc.
\newblock In \emph{Proceedings of Robotics: Science and Systems}, 2020.

\bibitem[Fisac et~al.(2019)Fisac, Lugovoy, Rubies-Royo, Ghosh, and
  Tomlin]{fisac2019bridging}
Jaime~F Fisac, Neil~F Lugovoy, Vicen{\c{c}} Rubies-Royo, Shromona Ghosh, and
  Claire~J Tomlin.
\newblock Bridging hamilton-jacobi safety analysis and reinforcement learning.
\newblock In \emph{2019 International Conference on Robotics and Automation
  (ICRA)}, pages 8550--8556, 2019.

\bibitem[Franklin et~al.(2002)Franklin, Powell, Emami-Naeini, and
  Powell]{franklin2002feedback}
Gene~F Franklin, J~David Powell, Abbas Emami-Naeini, and J~David Powell.
\newblock \emph{Feedback control of dynamic systems}, volume~4.
\newblock Prentice hall Upper Saddle River, NJ, 2002.

\bibitem[Gahlawat et~al.(2020)Gahlawat, Zhao, Patterson, Hovakimyan, and
  Theodorou]{gahlawat2020l1}
Aditya Gahlawat, Pan Zhao, Andrew Patterson, Naira Hovakimyan, and Evangelos
  Theodorou.
\newblock L1-gp: L1 adaptive control with bayesian learning.
\newblock In \emph{Learning for Dynamics and Control}, pages 826--837. PMLR,
  2020.

\bibitem[Gong et~al.(2019)Gong, Hartley, Da, Hereid, Harib, Huang, and
  Grizzle]{gong2019feedback}
Yukai Gong, Ross Hartley, Xingye Da, Ayonga Hereid, Omar Harib, Jiunn-Kai
  Huang, and Jessy Grizzle.
\newblock Feedback control of a cassie bipedal robot: Walking, standing, and
  riding a segway.
\newblock In \emph{2019 American Control Conference (ACC)}, pages 4559--4566,
  2019.

\bibitem[Grandia et~al.(2021)Grandia, Taylor, Ames, and
  Hutter]{grandia2021multi}
Ruben Grandia, Andrew~J Taylor, Aaron~D Ames, and Marco Hutter.
\newblock Multi-layered safety for legged robots via control barrier functions
  and model predictive control.
\newblock In \emph{2021 IEEE International Conference on Robotics and
  Automation (ICRA)}, pages 8352--8358, 2021.

\bibitem[He et~al.(2021)He, Zeng, Zhang, and Sreenath]{he2021rule}
Suiyi He, Jun Zeng, Bike Zhang, and Koushil Sreenath.
\newblock Rule-based safety-critical control design using control barrier
  functions with application to autonomous lane change.
\newblock In \emph{2021 American Control Conference (ACC)}, pages 178--185,
  2021.

\bibitem[Herbert et~al.(2017)Herbert, Chen, Han, Bansal, Fisac, and
  Tomlin]{herbert2017fastrack}
Sylvia~L Herbert, Mo~Chen, SooJean Han, Somil Bansal, Jaime~F Fisac, and
  Claire~J Tomlin.
\newblock Fastrack: A modular framework for fast and guaranteed safe motion
  planning.
\newblock In \emph{2017 IEEE 56th Annual Conference on Decision and Control
  (CDC)}, pages 1517--1522, 2017.

\bibitem[Hwangbo et~al.(2017)Hwangbo, Sa, Siegwart, and
  Hutter]{hwangbo2017control}
Jemin Hwangbo, Inkyu Sa, Roland Siegwart, and Marco Hutter.
\newblock Control of a quadrotor with reinforcement learning.
\newblock \emph{IEEE Robotics and Automation Letters}, 2\penalty0 (4):\penalty0
  2096--2103, 2017.

\bibitem[Ivanovic et~al.(2019)Ivanovic, Harrison, Sharma, Chen, and
  Pavone]{ivanovic2019barc}
Boris Ivanovic, James Harrison, Apoorva Sharma, Mo~Chen, and Marco Pavone.
\newblock Barc: Backward reachability curriculum for robotic reinforcement
  learning.
\newblock In \emph{2019 International Conference on Robotics and Automation
  (ICRA)}, pages 15--21, 2019.

\bibitem[Jin and Lavaei(2020)]{jin2020stability}
Ming Jin and Javad Lavaei.
\newblock Stability-certified reinforcement learning: A control-theoretic
  perspective.
\newblock \emph{IEEE Access}, 8:\penalty0 229086--229100, 2020.

\bibitem[Joshi and Chowdhary(2019)]{joshi2019deep}
Girish Joshi and Girish Chowdhary.
\newblock Deep model reference adaptive control.
\newblock In \emph{2019 IEEE 58th Conference on Decision and Control (CDC)},
  pages 4601--4608, 2019.

\bibitem[Kamthe and Deisenroth(2018)]{kamthe2018data}
Sanket Kamthe and Marc Deisenroth.
\newblock Data-efficient reinforcement learning with probabilistic model
  predictive control.
\newblock In \emph{International conference on artificial intelligence and
  statistics}, pages 1701--1710. PMLR, 2018.

\bibitem[Karl et~al.(2016)Karl, Soelch, Bayer, and Van~der Smagt]{karl2016deep}
Maximilian Karl, Maximilian Soelch, Justin Bayer, and Patrick Van~der Smagt.
\newblock Deep variational bayes filters: Unsupervised learning of state space
  models from raw data.
\newblock \emph{arXiv preprint arXiv:1605.06432}, 2016.

\bibitem[K{\"o}hler et~al.(2021)K{\"o}hler, K{\"o}tting, Soloperto,
  Allg{\"o}wer, and M{\"u}ller]{kohler2021robust}
Johannes K{\"o}hler, Peter K{\"o}tting, Raffaele Soloperto, Frank Allg{\"o}wer,
  and Matthias~A M{\"u}ller.
\newblock A robust adaptive model predictive control framework for nonlinear
  uncertain systems.
\newblock \emph{International Journal of Robust and Nonlinear Control},
  31\penalty0 (18):\penalty0 8725--8749, 2021.

\bibitem[Kolter and Manek(2019)]{manek2020learning}
J~Zico Kolter and Gaurav Manek.
\newblock Learning stable deep dynamics models.
\newblock \emph{Advances in neural information processing systems}, 32, 2019.

\bibitem[Lee et~al.(2020)Lee, Hwangbo, Wellhausen, Koltun, and
  Hutter]{lee2020learning}
Joonho Lee, Jemin Hwangbo, Lorenz Wellhausen, Vladlen Koltun, and Marco Hutter.
\newblock Learning quadrupedal locomotion over challenging terrain.
\newblock \emph{Science robotics}, 5\penalty0 (47), 2020.

\bibitem[Levine and Koltun(2013)]{levine2013guided}
Sergey Levine and Vladlen Koltun.
\newblock Guided policy search.
\newblock In \emph{International conference on machine learning}, pages 1--9.
  PMLR, 2013.

\bibitem[Levine et~al.(2016)Levine, Finn, Darrell, and Abbeel]{levine2016end}
Sergey Levine, Chelsea Finn, Trevor Darrell, and Pieter Abbeel.
\newblock End-to-end training of deep visuomotor policies.
\newblock \emph{The Journal of Machine Learning Research}, 17\penalty0
  (1):\penalty0 1334--1373, 2016.

\bibitem[Li et~al.(2020)Li, Cummings, and Sreenath]{li2020animated}
Zhongyu Li, Christine Cummings, and Koushil Sreenath.
\newblock Animated cassie: A dynamic relatable robotic character.
\newblock In \emph{2020 International Conference on Intelligent Robots and
  Systems (IROS)}, 2020.

\bibitem[Li et~al.(2021{\natexlab{a}})Li, Cheng, Peng, Abbeel, Levine, Berseth,
  and Sreenath]{li2021reinforcement}
Zhongyu Li, Xuxin Cheng, Xue~Bin Peng, Pieter Abbeel, Sergey Levine, Glen
  Berseth, and Koushil Sreenath.
\newblock Reinforcement learning for robust parameterized locomotion control of
  bipedal robots.
\newblock In \emph{2021 IEEE International Conference on Robotics and
  Automation (ICRA)}, pages 2811--2817, 2021{\natexlab{a}}.

\bibitem[Li et~al.(2021{\natexlab{b}})Li, Zeng, Chen, and
  Sreenath]{li2021vision}
Zhongyu Li, Jun Zeng, Shuxiao Chen, and Koushil Sreenath.
\newblock Vision-aided autonomous navigation of underactuated bipedal robots in
  height-constrained environments.
\newblock \emph{arXiv preprint arXiv:2109.05714}, 2021{\natexlab{b}}.

\bibitem[Ljung(1998)]{ljung1998system}
Lennart Ljung.
\newblock System identification.
\newblock In \emph{Signal analysis and prediction}, pages 163--173. Springer,
  1998.

\bibitem[Mania et~al.(2018)Mania, Guy, and Recht]{mania2018simple}
Horia Mania, Aurelia Guy, and Benjamin Recht.
\newblock Simple random search of static linear policies is competitive for
  reinforcement learning.
\newblock \emph{Advances in Neural Information Processing Systems}, 31, 2018.

\bibitem[Marvi and Kiumarsi(2021)]{marvi2021safe}
Zahra Marvi and Bahare Kiumarsi.
\newblock Safe reinforcement learning: A control barrier function optimization
  approach.
\newblock \emph{International Journal of Robust and Nonlinear Control},
  31\penalty0 (6):\penalty0 1923--1940, 2021.

\bibitem[Moldovan and Abbeel(2012)]{moldovan2012safe}
Teodor~Mihai Moldovan and Pieter Abbeel.
\newblock Safe exploration in markov decision processes.
\newblock In \emph{Proceedings of the 29th International Conference on Machine
  Learning (ICML)}, pages 1451--1458, 2012.

\bibitem[Nguyen et~al.(2018)Nguyen, Agrawal, Martin, Geyer, and
  Sreenath]{nguyen2018dynamic}
Quan Nguyen, Ayush Agrawal, William Martin, Hartmut Geyer, and Koushil
  Sreenath.
\newblock Dynamic bipedal locomotion over stochastic discrete terrain.
\newblock \emph{The International Journal of Robotics Research}, 37\penalty0
  (13-14):\penalty0 1537--1553, 2018.

\bibitem[Ostafew et~al.(2016)Ostafew, Schoellig, and
  Barfoot]{ostafew2016robust}
Chris~J Ostafew, Angela~P Schoellig, and Timothy~D Barfoot.
\newblock Robust constrained learning-based nmpc enabling reliable mobile robot
  path tracking.
\newblock \emph{The International Journal of Robotics Research}, 35\penalty0
  (13):\penalty0 1547--1563, 2016.

\bibitem[Peng et~al.(2018)Peng, Andrychowicz, Zaremba, and Abbeel]{peng2018sim}
Xue~Bin Peng, Marcin Andrychowicz, Wojciech Zaremba, and Pieter Abbeel.
\newblock Sim-to-real transfer of robotic control with dynamics randomization.
\newblock In \emph{2018 IEEE international conference on robotics and
  automation (ICRA)}, pages 3803--3810, 2018.

\bibitem[Peng et~al.(2020)Peng, Coumans, Zhang, Lee, Tan, and
  Levine]{peng2020learning}
Xue~Bin Peng, Erwin Coumans, Tingnan Zhang, Tsang-Wei~Edward Lee, Jie Tan, and
  Sergey Levine.
\newblock Learning agile robotic locomotion skills by imitating animals.
\newblock In \emph{Robotics: Science and Systems}, 07 2020.

\bibitem[Richards et~al.(2018)Richards, Berkenkamp, and
  Krause]{richards2018lyapunov}
Spencer~M Richards, Felix Berkenkamp, and Andreas Krause.
\newblock The lyapunov neural network: Adaptive stability certification for
  safe learning of dynamical systems.
\newblock In \emph{Conference on Robot Learning}, pages 466--476. PMLR, 2018.

\bibitem[Robey et~al.(2020)Robey, Hu, Lindemann, Zhang, Dimarogonas, Tu, and
  Matni]{robey2020learning}
Alexander Robey, Haimin Hu, Lars Lindemann, Hanwen Zhang, Dimos~V Dimarogonas,
  Stephen Tu, and Nikolai Matni.
\newblock Learning control barrier functions from expert demonstrations.
\newblock In \emph{2020 59th IEEE Conference on Decision and Control (CDC)},
  pages 3717--3724, 2020.

\bibitem[Robey et~al.(2021)Robey, Lindemann, Tu, and Matni]{robey2021learning}
Alexander Robey, Lars Lindemann, Stephen Tu, and Nikolai Matni.
\newblock Learning robust hybrid control barrier functions for uncertain
  systems.
\newblock \emph{IFAC-PapersOnLine}, 54\penalty0 (5):\penalty0 1--6, 2021.

\bibitem[Saveriano and Lee(2019)]{saveriano2019learning}
Matteo Saveriano and Dongheui Lee.
\newblock Learning barrier functions for constrained motion planning with
  dynamical systems.
\newblock In \emph{2019 IEEE/RSJ International Conference on Intelligent Robots
  and Systems (IROS)}, pages 112--119, 2019.

\bibitem[Schulman et~al.(2017)Schulman, Wolski, Dhariwal, Radford, and
  Klimov]{schulman2017proximal}
John Schulman, Filip Wolski, Prafulla Dhariwal, Alec Radford, and Oleg Klimov.
\newblock Proximal policy optimization algorithms.
\newblock \emph{arXiv preprint arXiv:1707.06347}, 2017.

\bibitem[Siekmann et~al.(2021)Siekmann, Godse, Fern, and
  Hurst]{siekmann2021sim}
Jonah Siekmann, Yesh Godse, Alan Fern, and Jonathan Hurst.
\newblock Sim-to-real learning of all common bipedal gaits via periodic reward
  composition.
\newblock In \emph{2021 IEEE International Conference on Robotics and
  Automation (ICRA)}, pages 7309--7315, 2021.

\bibitem[Singletary et~al.(2020)Singletary, Gurriet, Nilsson, and
  Ames]{singletary2020safety}
Andrew Singletary, Thomas Gurriet, Petter Nilsson, and Aaron~D. Ames.
\newblock Safety-critical rapid aerial exploration of unknown environments.
\newblock In \emph{2020 IEEE International Conference on Robotics and
  Automation (ICRA)}, pages 10270--10276, 2020.

\bibitem[Son and Nguyen(2019)]{son2019safety}
Tong~Duy Son and Quan Nguyen.
\newblock Safety-critical control for non-affine nonlinear systems with
  application on autonomous vehicle.
\newblock In \emph{2019 IEEE 58th Conference on Decision and Control (CDC)},
  pages 7623--7628, 2019.

\bibitem[Sui et~al.(2015)Sui, Gotovos, Burdick, and Krause]{sui2015safe}
Yanan Sui, Alkis Gotovos, Joel Burdick, and Andreas Krause.
\newblock Safe exploration for optimization with gaussian processes.
\newblock In \emph{International Conference on Machine Learning}, pages
  997--1005. PMLR, 2015.

\bibitem[Taylor et~al.(2020)Taylor, Singletary, Yue, and
  Ames]{taylor2020learning}
Andrew Taylor, Andrew Singletary, Yisong Yue, and Aaron Ames.
\newblock Learning for safety-critical control with control barrier functions.
\newblock In \emph{Learning for Dynamics and Control}, pages 708--717. PMLR,
  2020.

\bibitem[Todorov et~al.(2012)Todorov, Erez, and Tassa]{todorov2012mujoco}
Emanuel Todorov, Tom Erez, and Yuval Tassa.
\newblock Mujoco: A physics engine for model-based control.
\newblock In \emph{2012 IEEE/RSJ International Conference on Intelligent Robots
  and Systems}, pages 5026--5033, 2012.

\bibitem[Vinitsky et~al.(2020)Vinitsky, Du, Parvate, Jang, Abbeel, and
  Bayen]{vinitsky2020robust}
Eugene Vinitsky, Yuqing Du, Kanaad Parvate, Kathy Jang, Pieter Abbeel, and
  Alexandre Bayen.
\newblock Robust reinforcement learning using adversarial populations.
\newblock \emph{arXiv preprint arXiv:2008.01825}, 2020.

\bibitem[Wang et~al.(2021)Wang, Meng, Li, Smith, and Liu]{wang2021learning}
Chuanzheng Wang, Yiming Meng, Yinan Li, Stephen~L Smith, and Jun Liu.
\newblock Learning control barrier functions with high relative degree for
  safety-critical control.
\newblock In \emph{2021 European Control Conference (ECC)}, pages 1459--1464,
  2021.

\bibitem[Watter et~al.(2015)Watter, Springenberg, Boedecker, and
  Riedmiller]{watter2015embed}
Manuel Watter, Jost Springenberg, Joschka Boedecker, and Martin Riedmiller.
\newblock Embed to control: A locally linear latent dynamics model for control
  from raw images.
\newblock \emph{Advances in neural information processing systems}, 28, 2015.

\bibitem[Wei et~al.(2020)Wei, Kakade, and Ma]{wei2020implicit}
Colin Wei, Sham Kakade, and Tengyu Ma.
\newblock The implicit and explicit regularization effects of dropout.
\newblock In \emph{International Conference on Machine Learning}, pages
  10181--10192. PMLR, 2020.

\bibitem[Westenbroek et~al.(2020{\natexlab{a}})Westenbroek, Casta{\~n}eda,
  Agrawal, Sastry, and Sreenath]{westenbroek2020learning}
Tyler Westenbroek, Fernando Casta{\~n}eda, Ayush Agrawal, S~Shankar Sastry, and
  Koushil Sreenath.
\newblock Learning min-norm stabilizing control laws for systems with unknown
  dynamics.
\newblock In \emph{2020 59th IEEE Conference on Decision and Control (CDC)},
  pages 737--744, 2020{\natexlab{a}}.

\bibitem[Westenbroek et~al.(2020{\natexlab{b}})Westenbroek, Fridovich-Keil,
  Mazumdar, Arora, Prabhu, Sastry, and Tomlin]{westenbroek2020feedback}
Tyler Westenbroek, David Fridovich-Keil, Eric Mazumdar, Shreyas Arora, Valmik
  Prabhu, S~Shankar Sastry, and Claire~J Tomlin.
\newblock Feedback linearization for uncertain systems via reinforcement
  learning.
\newblock In \emph{2020 IEEE International Conference on Robotics and
  Automation (ICRA)}, pages 1364--1371, 2020{\natexlab{b}}.

\bibitem[Wright and Ma(2021)]{wright2021high}
John Wright and Yi~Ma.
\newblock \emph{High-dimensional data analysis with low-dimensional models:
  Principles, computation, and applications}.
\newblock Cambridge University Press, 2021.

\bibitem[Xie et~al.(2018)Xie, Berseth, Clary, Hurst, and van~de
  Panne]{xie2018feedback}
Zhaoming Xie, Glen Berseth, Patrick Clary, Jonathan Hurst, and Michiel van~de
  Panne.
\newblock Feedback control for cassie with deep reinforcement learning.
\newblock In \emph{2018 IEEE/RSJ International Conference on Intelligent Robots
  and Systems (IROS)}, pages 1241--1246, 2018.

\bibitem[Yaghoubi et~al.(2020)Yaghoubi, Fainekos, and
  Sankaranarayanan]{yaghoubi2020training}
Shakiba Yaghoubi, Georgios Fainekos, and Sriram Sankaranarayanan.
\newblock Training neural network controllers using control barrier functions
  in the presence of disturbances.
\newblock In \emph{2020 IEEE 23rd International Conference on Intelligent
  Transportation Systems (ITSC)}, pages 1--6, 2020.

\bibitem[Zeng et~al.(2021{\natexlab{a}})Zeng, Li, and
  Sreenath]{zeng2021enhancing}
Jun Zeng, Zhongyu Li, and Koushil Sreenath.
\newblock Enhancing feasibility and safety of nonlinear model predictive
  control with discrete-time control barrier functions.
\newblock In \emph{2021 60th IEEE Conference on Decision and Control (CDC)},
  pages 6137--6144, 2021{\natexlab{a}}.

\bibitem[Zeng et~al.(2021{\natexlab{b}})Zeng, Zhang, Li, and
  Sreenath]{zeng2021pointwise}
Jun Zeng, Bike Zhang, Zhongyu Li, and Koushil Sreenath.
\newblock Safety-critical control using optimal-decay control barrier function
  with guaranteed point-wise feasibility.
\newblock In \emph{2021 American Control Conference (ACC)}, pages 3856--3863,
  2021{\natexlab{b}}.

\bibitem[Zeng et~al.(2021{\natexlab{c}})Zeng, Zhang, and
  Sreenath]{zeng2021safety}
Jun Zeng, Bike Zhang, and Koushil Sreenath.
\newblock Safety-critical model predictive control with discrete-time control
  barrier function.
\newblock In \emph{American Control Conference (ACC)}, pages 3882--3889,
  2021{\natexlab{c}}.

\end{thebibliography}

\end{document}